%% file: arxiv.tex
\documentclass[sigconf]{acmart}

\setcopyright{none}
\renewcommand\footnotetextcopyrightpermission[1]{}

\definecolor{cfgreen}{HTML}{000000}
\definecolor{cfblue}{HTML}{000000}
\definecolor{cforange}{HTML}{000000}
\newcommand{\vbase}[1]{#1}
\newcommand{\vbaseB}[1]{\textbf{#1}}
\newcommand{\vbaseU}[1]{\underline{#1}}
\newcommand{\vstat}[1]{\textcolor{cfgreen}{#1}}
\newcommand{\vstatB}[1]{\textcolor{cfgreen}{\textbf{#1}}}
\newcommand{\vstatU}[1]{\textcolor{cfgreen}{\underline{#1}}}
\newcommand{\vmet}[1]{\textcolor{cfblue}{#1}}
\newcommand{\vmetB}[1]{\textcolor{cfblue}{\textbf{#1}}}
\newcommand{\vmetU}[1]{\textcolor{cfblue}{\underline{#1}}}
\newcommand{\vall}[1]{\textcolor{cforange}{#1}}
\newcommand{\vallB}[1]{\textcolor{cforange}{\textbf{#1}}}
\newcommand{\vallU}[1]{\textcolor{cforange}{\underline{#1}}}
\newcommand{\cfgbest}{}

\usepackage{longtable}
\usepackage{array}
\usepackage{pifont}
\usepackage{placeins}
\definecolor{tabgreen}{HTML}{2E7D32}
\definecolor{taborange}{HTML}{D97706}
\newcommand{\cmark}{\textcolor{tabgreen}{\ding{51}}}
\newcommand{\xmark}{\textcolor{black}{\ding{55}}}

\newcommand{\xmarknote}[1]{\makebox[1em][c]{\xmark}\smash{\rlap{\textsuperscript{#1}}}}

\makeatletter
\def\@subsubsecfont{\sffamily\bfseries\itshape}
\makeatother

\begin{document}

\title[UHI-Bench]{UHI-Bench: Benchmarking Dual-Source Urban Heat Island Modeling Across Cities in Diverse Climate Regimes}

\author{Wanyun Ling}
\affiliation{%
  \department{Department of Operations and Technology}
  \institution{Technical University of Munich}
  \city{Heilbronn}
  \country{Germany}}
\email{wanyun.ling@tum.de}

\author{Chenxi Liu}
\authornote{Corresponding authors.}
\affiliation{%
  \institution{Hong Kong Institute of Science \& Innovation, Chinese Academy of Sciences}
  \city{Hong Kong SAR}
  \country{China}}
\email{chenxi.liu@cair-cas.org.hk}

\author{Yi Xie}
\affiliation{%
  \department{Department of Operations and Technology}
  \institution{Technical University of Munich}
  \city{Heilbronn}
  \country{Germany}}
\email{aaron.xie@tum.de}

\author{Aopu Xu}
\affiliation{%
  \department{Faculty of Computer Science}
  \institution{RWTH Aachen University}
  \city{Aachen}
  \country{Germany}}
\email{aopu.xu@rwth-aachen.de}

\author{Zhuoqi Zeng}
\affiliation{%
  \institution{Hainan Bielefeld University of Applied Sciences}
  \city{Hainan}
  \country{China}}
\email{zhuoqi.zeng@hibiuh.edu.cn}

\author{Ziyue Li}
\authornotemark[1]
\affiliation{%
  \department{Department of Operations and Technology, Heilbronn Data Science Center, Munich Data Science Institute}
  \institution{Technical University of Munich}
  \city{Munich}
  \country{Germany}}
\email{ziyue.li@tum.de}

\renewcommand{\shortauthors}{Ling et al.}

\begin{abstract}
Urban heat islands (UHIs) are intensifying under climate change, exacerbating thermal exposure risks. Their two primary observations, land surface temperature UHI (LST-UHI) and near-surface air temperature UHI (AirT-UHI), capture physically distinct aspects of urban heat. However, most studies rely on a single source, and substituting one for the other can substantially bias the magnitude and spatial variability of human heat exposure. Accurate UHI modeling also requires dynamic meteorological drivers and static urban morphology features, but spatiotemporal incompatibilities hinder their alignment. Cloud gaps in LST observations and sparse AirT station networks further limit dual-source UHI modeling, motivating cross-city transfer across diverse climates. To bridge these gaps, we introduce \textbf{UHI-Bench}, the first UHI benchmark for dual-source UHI modeling that integrates dynamic and static environmental context. Following a unified signal, mechanism, and transfer framework, it evaluates over 20 baselines from four model families on five tasks across 20 cities and nine Köppen climate classes. Results show that no model is uniformly best, although foundation models remain consistently competitive and stable. Environmental covariates generally improve performance, but their utility varies across sources and tasks. Cross-city transferability is better explained by overlap in UHI regimes than by climate-zone similarity. With the dataset and standardized pipeline, our work provides practical guidance for urban heat modeling, promotes climate data equity, and supports future advances in climate research.
\end{abstract}




\maketitle

\begin{figure*}[!t]
  \centering
  \includegraphics[width=\textwidth]{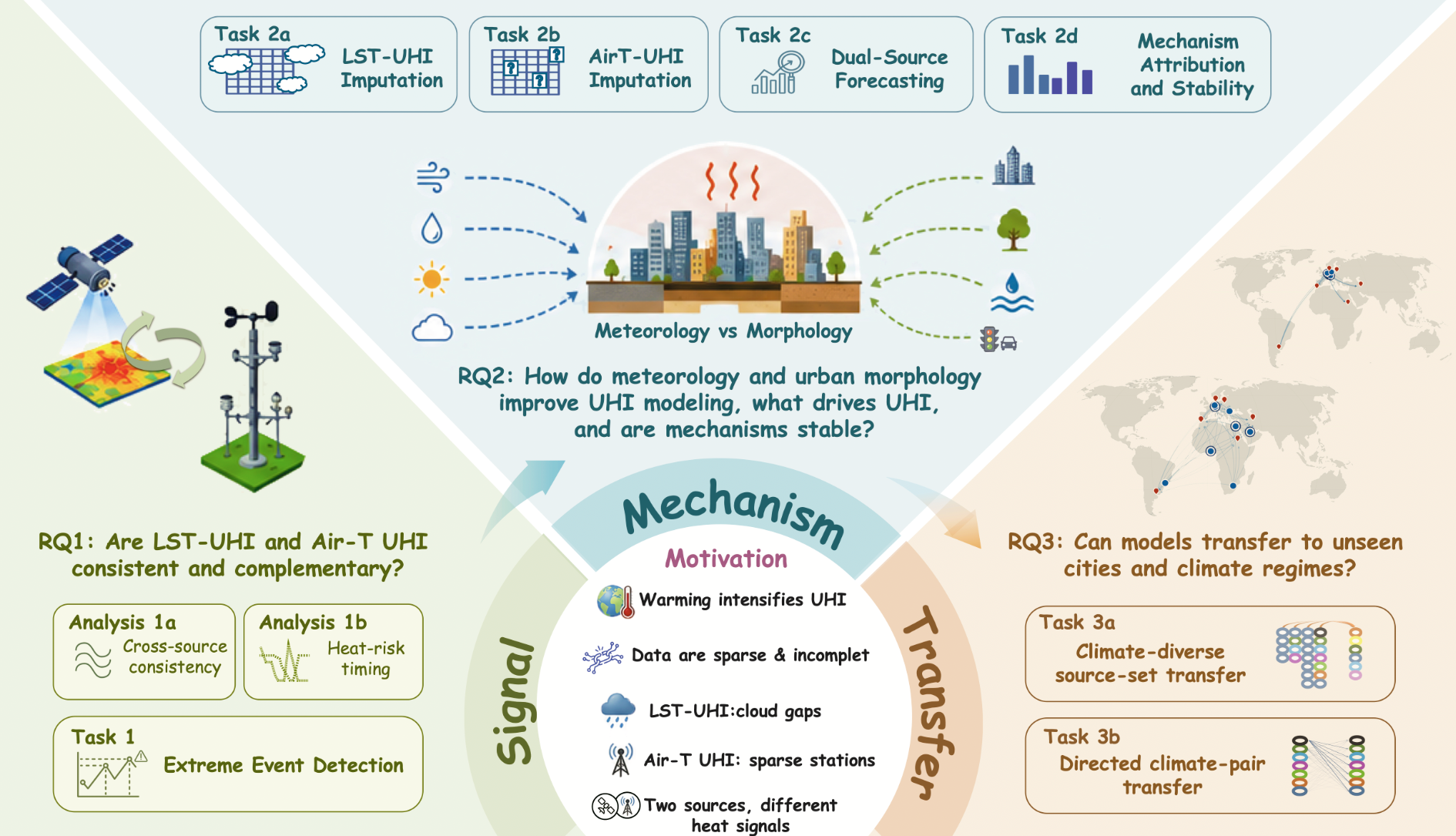}
  \caption{UHI-Bench benchmark overview. The benchmark is organized around
  three research questions, progressing from signal validity to mechanism
  understanding and finally to cross-city transfer.}
  \label{fig:benchmark-overview}
  \Description{Overview graphic summarizing the benchmark motivation,
  research questions, and tasks spanning signal consistency, detection,
  imputation, forecasting, attribution, and transfer.}
\end{figure*}

\section{Introduction}\label{sec:intro}\label{introduction}

Rapid urbanization and climate change are intensifying urban heat,
increasing thermal discomfort and heat-related health
risks~\cite{ref1,ref4,ref5,ref6,ref7}. In
metropolitan areas, this warming is often manifested as an urban heat
island (UHI) effect, defined as the temperature contrast between urban
areas and their surrounding rural or less-built reference
regions~\cite{ref12,ref23}. UHI is primarily observed through two
physically distinct signals: land-surface-temperature UHI (LST-UHI),
which captures radiative surface heating from satellite thermal
observations, and near-surface air-temperature UHI (AirT-UHI), which
more directly reflects pedestrian-level heat
exposure~\cite{ref8,ref9,ref10,ref11,ref12}.

Most existing UHI studies analyze only one observation source, and satellite-derived LST-UHI is frequently used as a substitute for near-surface air temperature because of its broader spatial availability~\cite{ref8,ref9,ref10}. Recent large-scale evidence shows that this substitution can produce substantial errors in both the estimated magnitude and intra-urban variability of human-relevant heat
exposure~\cite{ref56}. The two UHI signals respond differently to surface radiation, land-cover conditions, and atmospheric mixing~\cite{ref56,ref57,ref58}. Understanding these differences requires UHI observations to be analyzed jointly with both dynamic meteorological drivers and static urban-morphological features. However, spatiotemporally aligning these dynamic and static data remains a major challenge: their spatial and temporal resolutions are incompatible, meteorological products are often relatively coarse, and morphology datasets are updated only seasonally or annually~\cite{ref3,ref21,ref22,ref59}.

Data availability presents an additional challenge for dual-source UHI
modeling. LST-UHI has cloud-induced gaps and satellite sampling
constraints, whereas AirT-UHI requires dense, representative
urban station networks~\cite{ref11,ref12,ref60}. Many cities, particularly those in
observationally data-poor regions, lack such measurements across their
built environments~\cite{ref56,ref60}. These limitations raise a
central research question: can models transfer to unseen cities across
diverse climate regimes? The Local Climate Zone (LCZ) literature
suggests that geographically distant cities in different climate zones
may share similar morphology-related surface-energy and aerodynamic
characteristics~\cite{ref61}, while Demuzere et al.~\cite{ref55} show
that cross-city transfer may depend more on morphology, environmental
context, and input representations than on geographic proximity alone.
This provides a physically grounded precedent for investigating
cross-city urban-heat modeling. Advancing such cross-city transfer research could create new opportunities for climate studies, support sustainable urban adaptation, and contribute to more equitable urban-climate data and decision support.

To address these gaps, we introduce UHI-Bench, a benchmark for dual-source UHI modeling across cities in diverse climate regimes. UHI-Bench is built on UHI-Bench Data, which co-registers LST-UHI, AirT-UHI, hourly meteorological drivers, and static urban-morphology features on a standardized 1~km hourly grid.
The main dataset covers 20 cities across 9 K\"oppen climate classes~\cite{ref53}, including 16 cities with paired LST-UHI and AirT-UHI data from 2015 to 2025 and four additional LST-UHI-only cities from 2019 to 2025. Two supplementary cities provide station-format AirT observations. In total, the data comprise approximately 81,755 urban pixels. As illustrated in Figure~\ref{fig:benchmark-overview}, UHI-Bench follows a signal--mechanism--transfer framework, organized
around three research questions: dual-source signal validity (RQ1), environmental mechanism understanding (RQ2), and cross-city and cross-climate transfer (RQ3). Accordingly, it comprises two diagnostic analyses and five experimental tasks over four model groups: statistical methods, classical machine learning, deep spatiotemporal models, and time-series foundation models~\cite{ref75,ref65,ref66,ref73,ref74,ref68,ref71,ref67,ref72,ref69,ref70,ref63,ref62,ref64}.
\begin{table*}[!t]
  \caption{Comparison between UHI-Bench and related urban heat datasets.}
  \label{tab:dataset-comparison}
  \small
  \setlength{\tabcolsep}{2.5pt}
  \renewcommand{\arraystretch}{1.12}
  \resizebox{\textwidth}{!}{%
  \begin{tabular}{%
    >{\raggedright\arraybackslash}m{2.6cm}
    >{\raggedright\arraybackslash}m{1.8cm}
    >{\raggedright\arraybackslash}m{2.5cm}
    >{\centering\arraybackslash}m{1.2cm}
    >{\raggedright\arraybackslash}m{2.5cm}
    >{\raggedright\arraybackslash}m{1.4cm}
    >{\raggedright\arraybackslash}m{1.7cm}
    >{\centering\arraybackslash}m{1.7cm}
    >{\centering\arraybackslash}m{1.9cm}
    >{\centering\arraybackslash}m{1.2cm}
  }
    \toprule
    Dataset & Spatial coverage & Year span &
    UHI data available? & UHI source & Temporal resolution &
    Spatial resolution & Paired meteorological drivers &
    Paired static environmental data & Public access \\
    \midrule
    UrbClim~\cite{ref27}
      & 100 cities
      & 2008--2017
      & \xmark
      & Simulated AirT
      & Hourly
      & 100 m
      & \xmarknote{a}
      & \xmark
      & \cmark \\
    Global UHII~\cite{ref28}
      & 10{,}000+ cities
      & 2003--2020
      & \cmark
      & LST-UHI+AirT-UHI
      & Monthly
      & 1 km
      & \xmark
      & \xmark
      & \cmark \\
    Global HW Exposure~\cite{ref29}
      & Global
      & 2003--2020
      & \xmark
      & LST
      & Daily
      & 1 km
      & \xmark
      & \xmark
      & \cmark \\
    IBM Johannesburg~\cite{ref31}
      & 1 city
      & $\sim$10 years (not clearly specified)
      & \xmark
      & AirT
      & Daily
      & 1 km
      & \xmark
      & \xmark
      & \xmark \\
    US SUHI Database~\cite{ref32}
      & 497 US urbanized areas
      & 2013--2017
      & \cmark
      & LST
      & Seasonal
      & 1 km / census tract
      & \xmark
      & \xmarknote{b}
      & \xmarknote{e} \\
    SeoulTemp / DeepUHI~\cite{ref33}
      & 1 city: Seoul
      & 2021--2024
      & \xmark
      & AirT
      & Hourly
      & Street-level; 947 stations
      & \xmark
      & \xmarknote{c}
      & \cmark \\
    UHI-Bench
      & 22 cities
      & 2015--2025
      & \cmark
      & LST-UHI+AirT-UHI\textsuperscript{d}
      & Hourly
      & 1 km
      & \cmark
      & \cmark
      & \cmark \\
    \bottomrule
  \end{tabular}%
  }
  \vspace{2pt}
  \begin{minipage}{\textwidth}
    \scriptsize
    \textsuperscript{a} Meteorology-driven simulation, not paired
    ERA5-Land tensors.
    \textsuperscript{b} Limited static covariates, not paired gridded feature
    tensors.
    \textsuperscript{c} City-specific station inputs, not reusable 1~km
    morphology tensors.
    \textsuperscript{d} Dual-source co-registration applies to 16 core cities;
    the 22-city total includes 4 LST-only and 2 station-format AirT-only cities.
    \textsuperscript{e} Subscription-access paper; public GEE visualization
    available.
  \end{minipage}
\end{table*}
\enlargethispage{0\baselineskip}

Our main contributions are summarized as follows:
\par\vskip -3pt
\begin{itemize}
\setlength{\itemsep}{2pt plus 1pt minus 0.3pt}
\setlength{\parskip}{0pt plus 0.3pt}
\setlength{\parsep}{0pt plus 0.3pt}
\item
  To the best of our knowledge, UHI-Bench is the first UHI benchmark to support LST-UHI and AirT-UHI modeling.
\item
  Dynamic and static environmental data are integrated. UHI-Bench spatiotemporally aligns hourly meteorological drivers and urban-morphology features with dual-source UHI.
\item 
  A climate-aware benchmark for cross-city generalization. UHI-Bench defines a climate-diverse source-set and directed city-pair transfer protocols for evaluating generalization to unseen cities.
\item
  UHI-Bench evaluates over 20 baselines from 4 model families on 5 tasks across 20 cities and 9 K\"oppen climate classes. The results show that LST-UHI and AirT-UHI are complementary rather than interchangeable, environmental co-   variate utility varies across sources and tasks, and cross-city transferability is better explained by overlap in UHI regimes than by K\"oppen climate-label distance.
\end{itemize}

\vskip -4pt
\section{Related Work}\label{sec:related}\label{related-work}


Researchers have developed urban-temperature datasets from satellite
observations, weather stations, and climate products to characterize
UHI~\cite{ref10,ref12}. Early resources typically focused on a single source: satellite
products support broad LST-based surface-UHI analysis, whereas station
and gridded meteorological data provide AirT for canopy-layer studies~\cite{ref60}.
Because LST and AirT represent different physical layers, single-source
datasets cannot systematically compare LST-UHI and AirT-UHI.
Subsequent datasets improved spatial coverage and temporal resolution~\cite{ref59}.
UrbClim~\cite{ref27} provides hourly simulated AirT at 100 m resolution
for 100 European cities, but is limited to Europe and does not offer
paired UHI targets and aligned covariates. Global UHII~\cite{ref28}
covers more than 10,000 cities with monthly surface- and canopy-UHII
estimates, while the Global Heat Wave Exposure dataset~\cite{ref29}
provides daily 1 km MODIS-based heat-exposure estimates. Although
valuable for large-scale analysis, their temporal resolutions are too
coarse for hourly imputation, short-term forecasting, and
extreme-event detection, especially during rapidly evolving heat
episodes.

Other resources emphasize fine-grained
temperature mapping. Amplifier Air-Transformer~\cite{ref30} produces
hourly AirT maps over the contiguous United States, while IBM
Johannesburg studies~\cite{ref31}, the cross-city SUHII
dataset~\cite{ref32}, and SeoulTemp~\cite{ref33} support city- or
source-specific temperature modeling. However, these resources
generally provide absolute temperature or only one UHI source and do
not align LST-UHI and AirT-UHI across multiple climate zones.
Broader urban computing research has emphasized that urban and
environmental data are inherently heterogeneous and fragmented across
observation modalities, spatial resolutions, and data providers,
motivating systematic frameworks for cross-domain data fusion and
region-level representation learning across increasingly heterogeneous
urban settings~\cite{ref21,ref22}. Overall, existing datasets provide only subsets of the capabilities
required for comprehensive UHI benchmarking, as summarized in
Table~\ref{tab:dataset-comparison}. UHI-Bench addresses this gap by co-
registering dual-source UHI observations with hourly meteorological
variables and static urban features on a common 1 km grid across
multiple cities and climates. It supports the unified evaluation of the signal consistency, extreme detection, imputation, forecasting, attribution,
and cross-city transfer. Detailed related work on UHI modeling methods
and the remaining evaluation gaps are provided in
Appendix~\ref{sec:app-uhi-modeling}.

\begin{figure*}[!t]
  \centering
  \includegraphics[width=\textwidth]{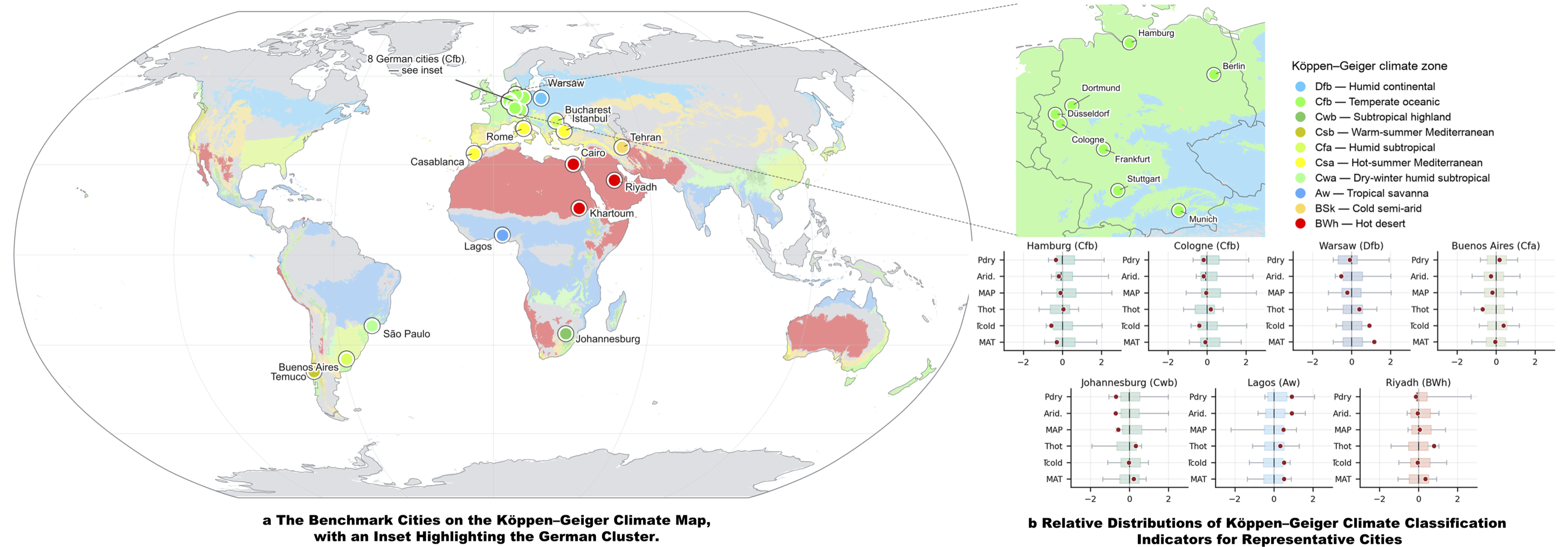}
  \caption{Geographical distribution and climate coverage of the
  22-city UHI-Bench Data.}
  \label{fig:uhi-city-distribution}
  \Description{Two-panel benchmark geography figure. Panel a maps the
  22 cities over global climate zones with a Germany inset. Panel b
  shows climate-indicator distributions for representative cities.}
\end{figure*}

\begin{figure*}[!t]
  \centering
  \includegraphics[width=0.9\textwidth]{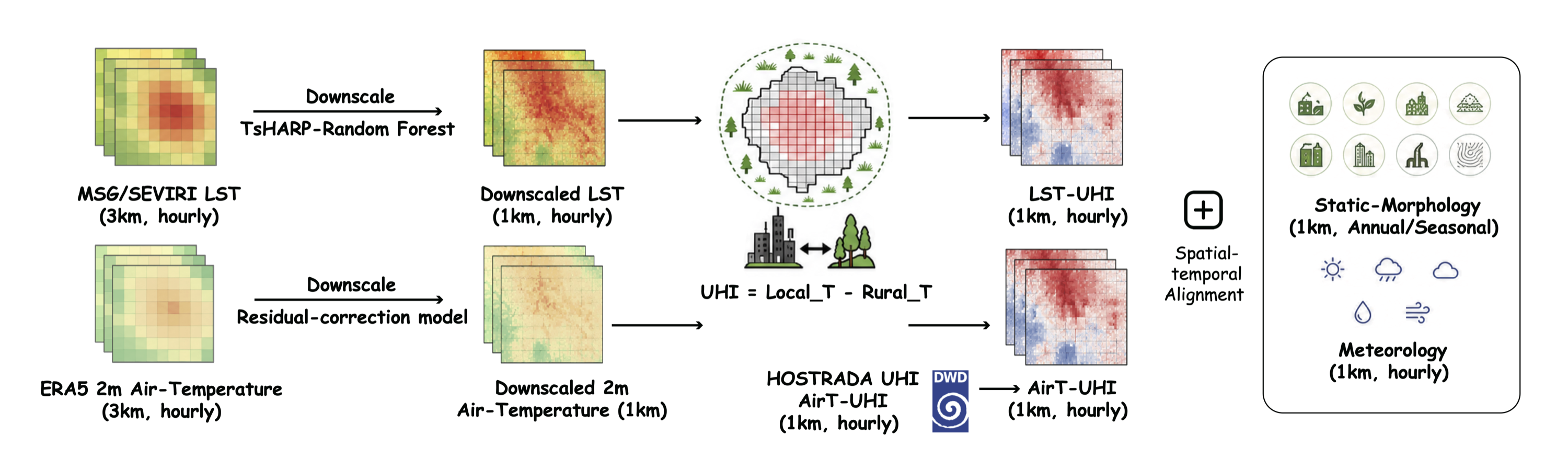}
  \caption{UHI-Bench data construction pipeline, including temperature
  downscaling, UHI derivation, and spatiotemporal alignment with urban
  morphology and meteorological variables.}
  \label{fig:construction-overview}
  \Description{Pipeline diagram showing downscaling of satellite land
  surface temperature and air temperature, UHI anomaly construction, and
  spatial-temporal alignment with static morphology and meteorological
  inputs.}
\end{figure*}

\section{UHI-Bench Data}\label{sec:bench}\label{uhi-bench}\label{overview}

As shown in Figure~\ref{fig:uhi-city-distribution}, UHI-Bench Data is
a multi-source, multi-city, and multi-climate-zone dataset for urban
heat island modeling. It covers 20 cities and two supplementary
station-based cities across 10 K\"oppen climate classes spanning four
continents and 15 countries, including tropical, arid, temperate, and
continental regimes.
UHI-Bench Data aligns LST-UHI, AirT-UHI, hourly meteorological
drivers, and static urban morphology features on a standardized 1 km
hourly pixel grid
\cite{ref11,ref37,ref37b,ref38,ref40,ref41,ref42,ref44,ref45,ref46,ref47,ref48,ref49,ref50,ref51,ref52,ref53}. The dataset includes 16 dual-source core
cities with
co-registered LST-UHI and AirT-UHI from 2015 to 2025, plus four
LST-UHI-only cities --- Tehran, Khartoum, Casablanca, and Istanbul ---
processed with the same LST, meteorological, and morphology pipelines
for surface-UHI transfer evaluation over 2019--2025, with two
supplementary cities providing station-format AirT observations. In total, the
dataset contains approximately 81,755 urban pixels, with detailed
city-level coverage, temporal ranges, and missing-data rates reported in
Appendix Table~\ref{tab:dataset_summary_full}. The specific computation of
the two UHI variables and their spatiotemporal alignment with
environmental context are described in Appendix~\ref{sec:data-processing}.

\subsection{Data Construction}\label{data-sources-and-construction}

\noindent As shown in Figure~\ref{fig:construction-overview}, the
dataset is constructed by aligning multi-source UHI observations,
meteorological drivers, and urban morphology features on a shared 1 km
hourly grid.

\subsubsection{LST-UHI}

LST-UHI is derived from the European Organisation for the Exploitation of Meteorological Satellites Satellite Application Facility
on Land Surface Analysis MSG/SEVIRI Land Surface Temperature
product, generated from observations acquired by the Spinning Enhanced
Visible and Infrared Imager (SEVIRI) onboard the Meteosat Second
Generation (MSG) geostationary satellites~\cite{ref37,ref37b}.
MSG/SEVIRI provides high-frequency thermal infrared observations and
covers all cities included in UHI-Bench, spanning Europe, Africa, the
Middle East, and parts of South America under a wide range of
atmospheric conditions.

The native MSG/SEVIRI LST product has a nominal spatial resolution of
approximately 3 km and is downscaled to 1 km using an RF-TsHARP
approach, consistent with random-forest-based regional LST downscaling
methods~\cite{ref25}. Downscaling quality is assessed through a 3 km round-trip
consistency, comparison with MODIS 1 km LST, and checks of expected
relationships with built-up coverage, NDVI, and water bodies. For German
cities, MSG/SEVIRI LST is aligned with the HOSTRADA 1 km grid.
LST-UHI is defined as local LST minus the contemporaneous mean LST of
surrounding rural reference pixels, selected primarily from cropland and
grassland using ESA WorldCover~\cite{ref50}. Cloud-contaminated observations are retained as missing
values and are not filled. The detailed downscaling method and
formulas are described in Appendix~\ref{sec:app-lst-uhi}.

\subsubsection{AirT-UHI}

For the eight German cities, AirT-UHI is obtained from the DWD
HOSTRADA gridded 2 m air-temperature UHI product~\cite{ref38}. For the eight
international cities, where comparable long-term, spatially continuous
AirT observations are unavailable, we construct an ERA5-constrained
1 km air-temperature downscaling product using a residual correction
model trained on HOSTRADA--ERA5 residuals from German cities with
meteorological, morphological, and temporal predictors, following
geographically-weighted-regression-based strategies for merging
coarse-resolution gridded products with ground observations~\cite{ref26,ref40}. We subsequently
compute AirT-UHI as the local air temperature minus the rural-ring mean
over a 15--25 km annulus, which also removes biases common to urban and
rural pixels. The residual model achieves held-out $R^2\approx0.79$,
indicating reliable reconstruction of local air-temperature variability.
The detailed construction procedure and formulas are described in
Appendix~\ref{sec:app-airt-uhi}.

\subsubsection{Meteorological and Urban Static Environmental Features}

Hourly meteorological variables are obtained from ERA5-Land and
resampled from the native grid to the standardized 1 km city grid~\cite{ref40,ref41}. The
variables include 10 m zonal and meridional wind ($u10$, $v10$), total
cloud cover ($tcc$), 2 m dewpoint temperature ($d2m$), boundary-layer
height ($blh$), and surface solar radiation downwards ($ssrd$). Urban
static features are compiled from open geospatial sources and include
building coverage, road and POI density, nighttime light, NDVI, water
fraction, distance to water, elevation, building height, and wind
exposure, capturing complementary aspects of surface cover, human
activity, and terrain that shape local heat retention and airflow.
These features have been widely recognized as key determinants
of AirT-UHI and LST-UHI variability and heat-exposure disparities in
urban regions~\cite{ref34,ref35,ref36,ref42,ref44,ref45,ref46,ref47,ref48,ref49,ref50,ref54}. Annual or seasonal versions are provided
where available, to capture longer-term morphological and phenological
change. Full feature definitions and processing are described in
Appendix Tables~\ref{tab:feature-formulas}, \ref{tab:glossary-static}, and
\ref{tab:glossary-meteo}.
\vskip -4pt
\subsection{Data Splits, Reproducibility, and Extensibility}\label{data-split-reproducibility-and-extensibility}

The default split uses 2015--2022 for training and statistics and
2023--2025 for evaluation. The four LST-UHI-only cities cover
2019--2025 and use the same 2023--2025 evaluation period. Task-specific
rules are described in Section~\ref{experimental-setup} and
Appendix Table~\ref{tab:task-overview}.

\textbf{UHI-Bench is released with the construction pipeline, standardized
dataloaders, and baseline implementations. The code repository\footnote{\url{https://github.com/wyling0v0/UHI-Bench.git}} and the dataset \footnote{\url{https://huggingface.co/datasets/WyLing0v0/uhi-bench}.}} are available. The workflow can
be extended to other cities within the MSG/SEVIRI full-disk domain using
the same LST downscaling, rural-reference differencing, and
feature-construction procedures.

\section{Experiments}\label{sec:exp}\label{experiments}

In this section, we conduct extensive experiments to investigate three
climate-aware research questions, progressing from signal validity, to
mechanism understanding, and finally to cross-climate transfer:

\begin{itemize}
\setlength{\itemsep}{1pt}
\setlength{\parsep}{0pt}
\setlength{\topsep}{2pt}
\setlength{\partopsep}{0pt}
\item
  RQ1: Are \textit{LST-UHI} and \textit{AirT-UHI} consistent and
  complementary indicators of urban heat risk across different climate
  zones?
\item
  RQ2: How do \textit{dynamic meteorology} and \textit{static urban
  morphology} contribute to UHI modeling, and are their predictive
  utility and inferred drivers stable across climate zones?
\item
  RQ3: Can models generalize to \textit{unseen cities} and climate
  regimes, is source-climate diversity more important than simply
  increasing the number of source cities, and can cities from
  \textit{different climate regimes} transfer to each other?
\end{itemize}

\subsection{Experimental Setup}\label{experimental-setup}
We evaluate UHI-Bench through a signal--mechanism--transfer protocol:
dual-source consistency and extreme-event detection, imputation and
forecasting with dynamic and static covariates, and cross-city transfer
under climate-diverse source settings. All tasks follow the temporal
splits and leakage-control protocol in Section~3.3. Detailed task
definitions, baseline groups, input configurations, city coverage, and
metrics are provided in Appendix~\ref{sec:app-experimental-setup} and
Appendix Tables~\ref{tab:baseline-groups} and \ref{tab:task-overview}.

\vskip -8pt
\subsection{Dual-source Consistency (RQ1)}\label{dual-source-consistency-and-heat-risk-detection}

\subsubsection{Cross-source Consistency and Complementarity (Analysis~1a)}\label{cross-source-consistency-and-complementarity}

\begin{figure}[!tbp]
  \centering
  \includegraphics[width=\linewidth]{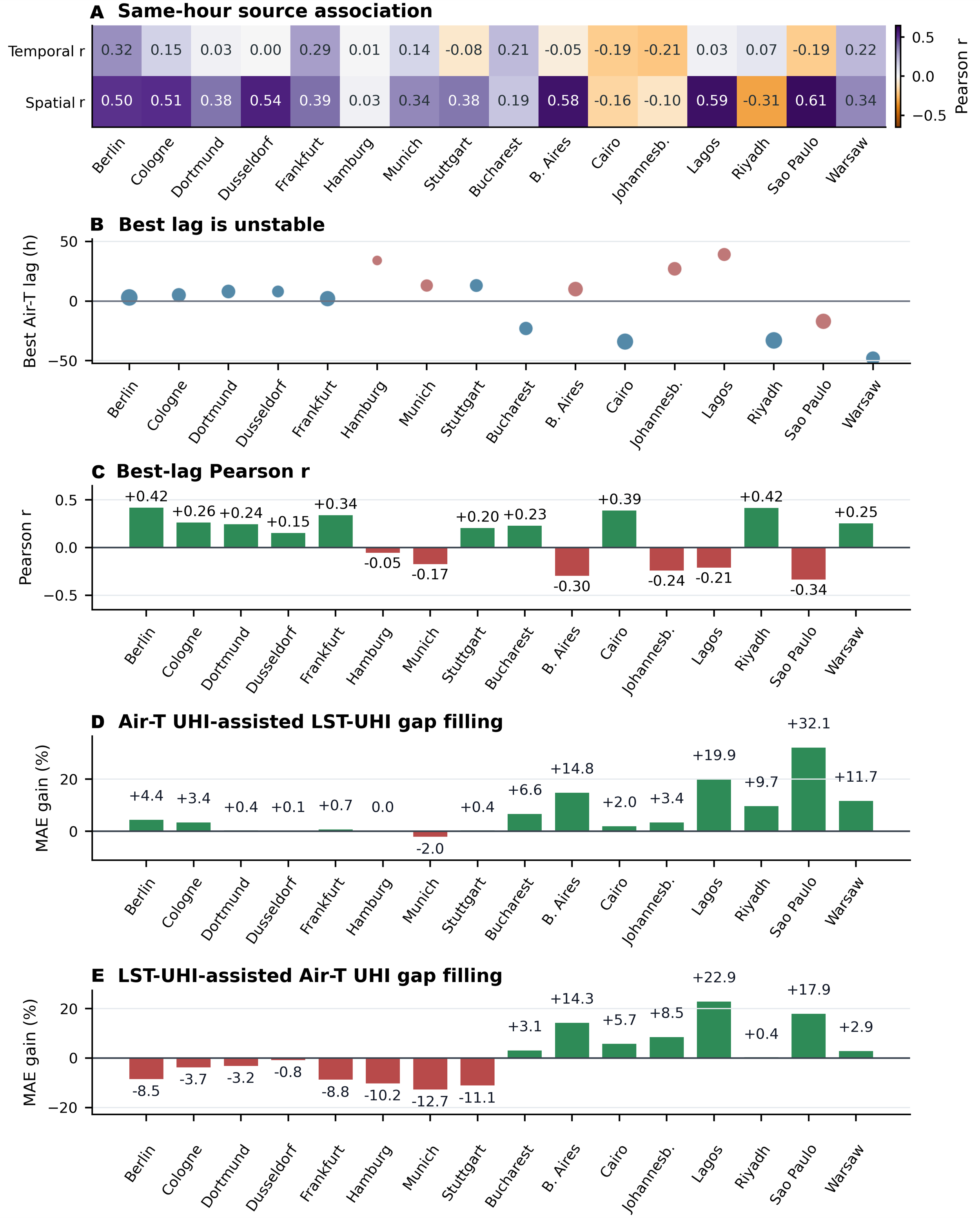}
  \caption{Analysis~1a cross-source consistency across 16 cities.}
  \label{fig:cross-source-summary}
  \Description{Cross-source diagnostic summary with same-hour correlation,
  lag instability, best-lag Pearson correlation, and gap-filling gain panels.}
\end{figure}

Figure~\ref{fig:cross-source-summary} shows
that \textbf{LST-UHI and AirT-UHI are related but not interchangeable.}
Same-hour temporal correlations (A) are generally weak, whereas spatial
correlations are relatively stronger across the sixteen paired cities,
indicating that the
two sources agree more on broad urban heat patterns than on hour-to-hour
dynamics. The best-lag analysis (B) further shows that lag correction
cannot reliably align the two sources: optimal lags vary widely across
cities, and best-lag correlations (C) remain weak or negative in several
cases.
Cross-source imputation confirms this asymmetric relationship. AirT-UHI
provides limited but useful support for LST-UHI imputation,
especially in several non-German cities such as Buenos Aires, Lagos,
Riyadh, São Paulo, and Warsaw. In contrast, LST-UHI does not
consistently improve sparse AirT-UHI imputation, and often hurts
performance in German cities. This is likely because German AirT fields
are already easier to interpolate, and because the temperate oceanic climate (Cfb), with
frequent cloud cover, stronger wind, and boundary-layer mixing can
decouple surface temperature from near-surface air temperature.

\subsubsection{Heat-risk Timing and Extreme-event Detection (Analysis~1b, Task~1)}

\begin{figure}[t]
  \centering
  \includegraphics[width=\linewidth]{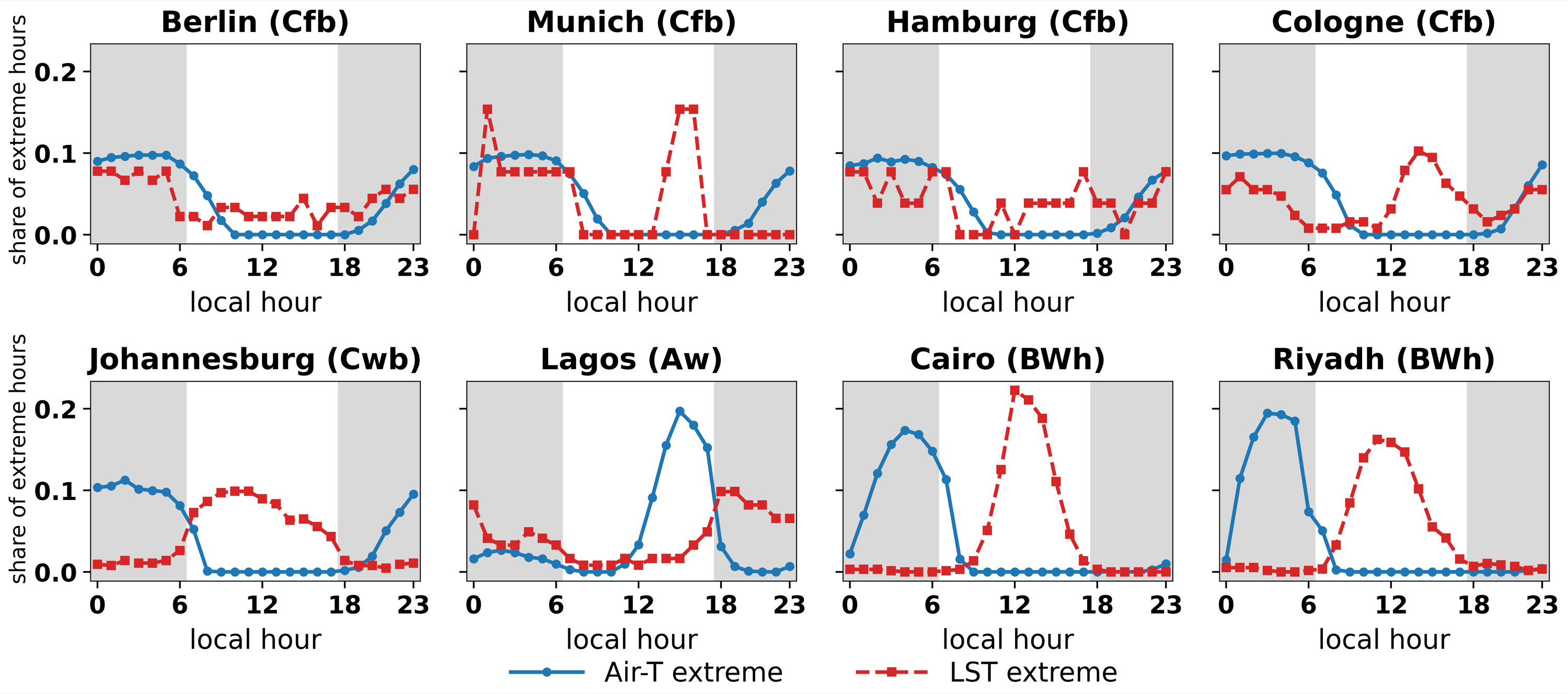}
  \caption{Analysis~1b extreme UHI event detection by hour of day and city.}
  \label{fig:diurnal-extremes}
  \Description{City-wise diurnal profiles of AirT and LST extreme-hour shares
  across the benchmark cities.}
\end{figure}

Analysis 1b
heat-risk timing diagnostics show that \textbf{LST-UHI and AirT-UHI
capture different heat-risk timing.} In Figure~\ref{fig:diurnal-extremes}, the y-axis reports the share of
all extreme hours assigned to each hour of the day, rather than the
probability of an extreme event at that
hour. Even so, the two sources'
extremes are not always synchronized. In the Cfb cities, AirT extremes tend
to concentrate in nighttime or early-morning hours, whereas LST extremes
are sparser and more often shifted toward daytime or fragmented peaks.
In warmer climates, the phase mismatch is even more pronounced.
Cairo and Riyadh show a stronger contrast between nocturnal AirT
extremes and daytime LST extremes.

Table~\ref{tab:app1c} further shows that \textbf{AirT-UHI extremes are easier to
detect than LST-UHI extremes.} For extreme-event detection,
the ML model XGBoost performs best for AirT-UHI, whereas the foundation
model Chronos-2 performs best for LST-UHI. For both UHI sources, adding
meteorological features improves detection performance. This matches the
smoother AirT signal and the sharper, cloud-sensitive LST signal.
Complete results for Berlin, Munich, Hamburg, and the remaining
baselines, including Unsupervised ML and DL, are reported in Appendix
Table~\ref{tab:1c-fm-complete}.

\vskip -2pt
\vskip -4pt

\begin{table}[t]
\setlength{\abovecaptionskip}{2pt}
\setlength{\belowcaptionskip}{2pt}
\caption{Task~1 detection leaderboard for Lagos, Johannesburg, Cologne, and Riyadh.}
\label{tab:app1c}
\normalsize
\setlength{\tabcolsep}{1pt}
\renewcommand{\arraystretch}{1.0}
\resizebox{\columnwidth}{!}{%
\begin{tabular}{@{}l l c c c c@{}}
\toprule
& & \multicolumn{4}{c}{F1 (AirT-UHI|LST-UHI)} \\
\cmidrule(lr){3-6}
Group & Model & Lagos (Aw) & Johannesburg (Cwb) & Cologne (Cfb) &
Riyadh (BWh) \\
\midrule
Stat/Geo & Percentile & \vbase{0.654}|\vbase{0.248} &
\vbase{0.760}|\vbase{0.621} & \vbase{0.875}|\vbase{0.268} &
\vbase{0.786}|\vbase{0.329} \\

Stat/Geo & SeasonalNaive & \vbase{0.371}|\vbase{0.249} &
\vbase{0.508}|\vbase{0.540} & \vbase{0.496}|\vbase{0.231} &
\vbase{0.462}|\vbase{0.187} \\

\midrule

ML & RF & \vbase{0.740}|\vbase{0.378} &
\vbase{0.739}|\vbase{0.663} & \vbaseU{0.939}|\vbase{0.261} &
\vbase{0.870}|\vbase{0.481} \\

ML & XGBoost & \vbase{0.726}|\vbase{0.361} &
\vbase{0.713}|\vbase{0.633} & \vbase{0.938}|\vbase{0.226} &
\vbaseU{0.876}|\vbase{0.460} \\

ML & RF+m & \vmet{0.752}|\vmet{0.396} &
\vmet{0.753}|\vmet{0.659} & \vmet{0.923}|\vmet{0.294} &
\vmet{0.872}|\vmetU{0.506} \\

ML & XGBoost+m &
\cfgbest\vmetB{0.789}|\vmet{0.377} &
\cfgbest\vmetB{0.786}|\vmet{0.672} &
\cfgbest\vmetB{0.949}|\vmet{0.308} &
\cfgbest\vmetB{0.898}|\vmet{0.462} \\

\midrule

FM & Chronos-2 & \vbase{0.754}|\vbaseU{0.441} &
\vbase{0.773}|\vbaseU{0.689} & \vbase{0.926}|\vbaseU{0.423} &
\vbase{0.857}|\vbase{0.483} \\

FM & TimesFM & \vbase{0.747}|\vbase{0.303} &
\vbase{0.748}|\vbase{0.666} & \vbase{0.909}|\vbase{0.315} &
\vbase{0.846}|\vbase{0.422} \\

FM & MOIRAI-2+m & \vmet{0.665}|\vmet{0.327} &
\vmet{0.734}|\vmet{0.653} & \vmet{0.933}|\vmet{0.339} &
\vmet{0.780}|\vmet{0.376} \\

FM & Chronos-2+m & \vmetU{0.760}|\cfgbest\vmetB{0.501} &
\vmetU{0.777}|\cfgbest\vmetB{0.708} &
\vmet{0.916}|\cfgbest\vmetB{0.471} &
\vmet{0.862}|\cfgbest\vmetB{0.567}
\\

\bottomrule

\end{tabular}%

}

\vspace{2pt}
\parbox{\columnwidth}{\scriptsize OrdKrig/RegKrig = Ordinary/Regression
Kriging; GWNet = GraphWaveNet; IGNNK = GNN (IGNNK); RF = Random Forest.}

\end{table}

\subsection{Mechanisms and Covariate Utility (RQ2)}\label{operational-uhi-modeling-and-mechanism-analysis}

\subsubsection{LST-UHI Cloud-gap Imputation (Task~2a)}
Table~\ref{tab:1a} shows that \textbf{the foundation model
Chronos-2 achieves the best overall performance when meteorological
features are included, yielding the lowest average MAE. The performance
gains are most evident in Cairo and Lagos.} Results are grouped by
cloud-cover missing-\% bin; ML baseline results for RF and
XGBoost are reported in Appendix Table~\ref{tab:task2a-remaining}.

GraphWaveNet is the strongest non-foundation baseline under large cloud
gaps, indicating that temporal context and spatial dependencies are
critical for imputing contiguous missing regions. With real cloud
masks, imputation difficulty depends not only on the missing
fraction but also on the spatial variability of the obscured areas,
which explains why Bucharest can be more challenging than Lagos despite
its lower overall missingness. Foundation-model gains in
high-missingness cases should therefore be interpreted together with the
spatial structure of the missing region. Covariate utility is also
model- and climate-dependent: meteorological inputs are particularly
beneficial in hot-arid and tropical cities such as Cairo and Lagos,
where surface temperature is strongly affected by radiation, cloud
cover, boundary-layer conditions, and wind. In Bucharest, by contrast,
history-only foundation models are already competitive, while static
urban features mainly improve classical spatial models and do not
consistently benefit foundation models.

\begin{table}[t]

\caption{Task~2a LST cloud-gap imputation MAE for Cairo, Bucharest, and Lagos.}\label{tab:1a}

\small

\setlength{\tabcolsep}{0.5pt}

\resizebox{\columnwidth}{!}{%
\begin{tabular}{@{}l l c c c c c c c c c c c c@{}}

\toprule

& & \multicolumn{12}{c}{MAE} \\

\cmidrule(lr){3-14}

& & \multicolumn{4}{c}{Cairo (BWh,
20.0\%)} &
\multicolumn{4}{c}{Bucharest (Cfb,
57.2\%)} & \multicolumn{4}{c}{Lagos
(Aw, 75.6\%)} \\

\cmidrule(lr){3-6}\cmidrule(lr){7-10}\cmidrule(lr){11-14}

Group & Method & 0--25\% & 25--50\%
& 50--75\% & >75\% &
0--25\% & 25--50\% &
50--75\% & >75\% &
0--25\% & 25--50\% &
50--75\% & >75\%
\\

\midrule

Stat/Geo & IDW & \vbase{1.405} &
\vbase{1.568} & \vbase{2.082} &
\vbase{2.258} & \vbase{0.989} &
\vbase{1.379} & \vbase{1.285} &
\vbase{1.440} & \vbase{1.500} &
\vbase{1.656} & \vbase{2.005} &
\vbase{2.521} \\

Stat/Geo & OrdKrig & \vbase{1.440} &
\vbase{1.623} & \vbase{2.172} &
\vbase{2.311} & \vbase{1.007} &
\vbase{1.414} & \vbase{1.346} &
\vbase{1.471} & \vbase{1.767} &
\vbase{1.838} & \vbase{2.161} &
\vbase{2.599} \\

Stat/Geo & RegKrig+s & \vstat{1.255}
& \vstat{1.216} & \vstat{1.534} &
\vstat{1.746} & \vstat{0.969} &
\vstat{1.262} & \vstat{1.205} &
\vstat{1.350} & \vstat{1.052} &
\vstat{1.125} & \vstat{1.244} &
\vstat{1.397} \\

\midrule

DL & SAITS & \vbase{1.077} &
\vbase{1.124} & \vbase{1.071} &
\vbase{1.099} & \vbase{1.843} &
\vbase{1.844} & \vbase{1.797} &
\vbase{1.818} & \vbase{1.223} &
\vbase{1.302} & \vbase{1.355} &
\vbase{1.337} \\

DL & GWNet & \vbaseU{0.548} &
\vbase{0.569} & \vbase{0.628} &
\vbase{0.678} &
\cfgbest\vbaseB{0.656} &
\vbase{0.666} & \vbase{0.715} &
\vbase{1.007} & \vbase{0.693} &
\vbase{0.701} & \vbase{0.778} &
\vbase{0.844} \\

DL & GWNet+m &
\cfgbest\vmetB{0.542} &
\vmetU{0.563} & \vmetU{0.566} &
\vmet{0.579} & \vmet{1.069} &
\vmet{0.968} & \vmet{1.001} &
\vmet{0.997} & \vmet{0.702} &
\vmet{0.706} & \vmet{0.774} &
\vmet{0.846} \\

DL & IGNNK+s & \vstat{1.754} &
\vstat{1.927} & \vstat{1.958} &
\vstat{2.098} & \vstat{1.489} &
\vstat{1.856} & \vstat{1.824} &
\vstat{1.997} & \vstat{1.466} &
\vstat{1.659} & \vstat{1.744} &
\vstat{1.864} \\

DL & GWNet+s & \vstat{0.551} &
\vstat{0.579} & \vstat{0.605} &
\vstat{0.639} & \vstatU{0.666} &
\vstat{0.683} & \vstat{0.704} &
\vstat{0.770} & \vstat{0.721} &
\vstat{0.731} & \vstat{0.818} &
\vstat{0.946} \\

\midrule

FM & TimesFM & \vbase{0.646} &
\vbase{0.606} & \vbase{0.596} &
\vbase{0.592} & \vbase{0.689} &
\vbaseU{0.632} & \vbaseU{0.644} &
\vbaseU{0.637} & \vbaseU{0.593} &
\vbaseU{0.614} & \vbaseU{0.594} &
\vbaseU{0.588} \\

FM & Chronos-2 & \vbase{0.626} &
\vbase{0.583} & \vbase{0.579} &
\vbaseU{0.572} & \vbase{0.675} &
\cfgbest\vbaseB{0.627} &
\cfgbest\vbaseB{0.638} &
\cfgbest\vbaseB{0.630} &
\vbase{0.599} & \vbase{0.619} &
\vbase{0.598} & \vbase{0.590}
\\

FM & Chronos-2+m & \vmet{0.590} &
\cfgbest\vmetB{0.556} &
\cfgbest\vmetB{0.547} &
\cfgbest\vmetB{0.541} &
\vmet{0.716} & \vmet{0.667} &
\vmet{0.682} & \vmet{0.672} &
\cfgbest\vmetB{0.587} &
\cfgbest\vmetB{0.607} &
\cfgbest\vmetB{0.589} &
\cfgbest\vmetB{0.581}
\\

\bottomrule

\end{tabular}%
}

\end{table}

\subsubsection{AirT-UHI Sparse Imputation (Task~2b)}
Table~\ref{tab:1b} shows that \textbf{AirT-UHI sparse imputation is generally easier than LST-UHI cloud-gap imputation because AirT fields are smoother and the missing pixels are randomly scattered.} Classical
machine-learning baselines are already competitive under low to moderate
missing rates. Results are grouped by station-missing bin; full
Lagos/Cologne/Riyadh results are reported in Appendix
Table~\ref{tab:task2b-table4-continuation}, and Munich/Cairo results
are reported in Appendix Table~\ref{tab:task2b-complete}.

Model behavior differs across climate regimes. In tropical Lagos,
XGBoost with static urban morphology features achieves the best
performance across all missing-rate bins, suggesting that urban-form covariates capture much of the persistent
AirT-UHI spatial pattern. In the Cfb cities, represented by Cologne in
the main table and Munich, temporal continuity is more
important. The foundation model Chronos-2 remains
highly stable even without auxiliary features, while the DL model
GraphWaveNet performs well when both static and meteorological features
are included. This pattern indicates that AirT-UHI fields in the Cfb cities are smoother and more temporally
constrained, so auxiliary features help selectively rather than
universally. In hot-desert Riyadh, GraphWaveNet
with both static and meteorological environmental features performs best
across all missing-rate bins, showing that desert AirT imputation
requires both spatial structure and dynamic weather context. These results reveal a source-dependent covariate pattern in UHI
imputation. LST-UHI cloud-gap imputation benefits more from
meteorological environmental features. In contrast, AirT-UHI sparse
imputation often benefits more from static urban morphology.

\vskip -2pt
\begin{table}[t]

\caption{Task~2b AirT station-sparse imputation MAE for Lagos, Cologne, and Riyadh.}\label{tab:1b}
\vspace{-3pt}

\small

\setlength{\tabcolsep}{0.5pt}

  \resizebox{\columnwidth}{!}{%

\begin{tabular}{@{}l l c c c c c c c c c c c c@{}}

\toprule

& & \multicolumn{12}{c}{MAE, by station
missing-\% bin} \\

\cmidrule(lr){3-14}

& & \multicolumn{4}{c}{Lagos (Aw)} &
\multicolumn{4}{c}{Cologne (Cfb)} &
\multicolumn{4}{c}{Riyadh (BWh)}
\\

\cmidrule(lr){3-6}\cmidrule(lr){7-10}\cmidrule(lr){11-14}

Group & Method & 0--25\% & 25--50\%
& 50--75\% & >75\% &
0--25\% & 25--50\% &
50--75\% & >75\% &
0--25\% & 25--50\% &
50--75\% & >75\%
\\

\midrule

ML & RF & \vbase{0.052} &
\vbase{0.053} & \vbase{0.054} &
\vbase{0.068} & \vbase{0.1104} &
\vbase{0.1188} & \vbase{0.1310} &
\vbase{0.1602} & \vbase{0.0776} &
\vbase{0.0772} & \vbase{0.0792} &
\vbase{0.0933} \\

ML & XGBoost & \vbase{0.053} &
\vbase{0.054} & \vbase{0.055} &
\vbase{0.068} & \vbase{0.1061} &
\vbase{0.1147} & \vbase{0.1287} &
\vbase{0.1595} & \vbase{0.0784} &
\vbase{0.0781} & \vbase{0.0796} &
\vbase{0.0952} \\

ML & RF+s & \vstatU{0.030} &
\vstatU{0.030} & \vstatU{0.031} &
\vstatU{0.038} & \vstat{0.0951} &
\vstat{0.0995} & \vstat{0.1066} &
\vstat{0.1200} & \vstat{0.0525} &
\vstat{0.0524} & \vstat{0.0527} &
\vstat{0.0644} \\

ML & XGBoost+s &
\cfgbest\vstatB{0.027} &
\cfgbest\vstatB{0.027} &
\cfgbest\vstatB{0.027} &
\cfgbest\vstatB{0.034} &
\vstat{0.0863} & \vstat{0.0924} &
\vstat{0.1017} & \vstat{0.1191} &
\vstat{0.0484} & \vstat{0.0483} &
\vstat{0.0488} & \vstatU{0.0590}
\\

\midrule

DL & SAITS & \vbase{0.097} &
\vbase{0.096} & \vbase{0.096} &
\vbase{0.095} & \vbase{0.2270} &
\vbase{0.2302} & \vbase{0.2280} &
\vbase{0.2296} & \vbase{0.2170} &
\vbase{0.2136} & \vbase{0.2152} &
\vbase{0.2140} \\

DL & GWNet & \vbase{0.0539} &
\vbase{0.0558} & \vbase{0.0624} &
\vbase{0.0916} & \vbase{0.0292} &
\vbaseU{0.0287} & \vbase{0.0408} &
\vbase{0.0973} & \vbase{0.0330} &
\vbase{0.0348} & \vbaseU{0.0391} &
\vbase{0.1143} \\

DL & GWNet+m & \vmet{0.054} &
\vmet{0.057} & \vmet{0.063} &
\vmet{0.092} & \vmet{0.0283} &
\vmet{0.0343} & \vmet{0.0478} &
\vmet{0.0875} & \vmet{0.0329} &
\vmet{0.0343} & \vmet{0.0928} &
\vmet{0.2923} \\

DL & GWNet+s+m & \vall{0.053} &
\vall{0.055} & \vall{0.063} &
\vall{0.105} &
\cfgbest\vallB{0.0233} &
\cfgbest\vallB{0.0251} &
\vall{0.0380} & \vall{0.0722} &
\cfgbest\vallB{0.0305} &
\cfgbest\vallB{0.0309} &
\cfgbest\vallB{0.0353} &
\cfgbest\vallB{0.0562}
\\

DL & IGNNK+s & \vstat{0.049} &
\vstat{0.051} & \vstat{0.063} &
\vstat{0.105} & \vstat{0.1051} &
\vstat{0.1003} & \vstat{0.1168} &
\vstat{0.1900} & \vstat{0.0699} &
\vstat{0.0716} & \vstat{0.0896} &
\vstat{0.1795} \\

DL & GWNet+s & \vstat{0.0515} &
\vstat{0.0550} & \vstat{0.0668} &
\vstat{0.1161} & \vstatU{0.0254} &
\vstat{0.0287} & \vstat{0.0345} &
\vstat{0.0697} & \vstatU{0.0317} &
\vstatU{0.0328} & \vstat{0.0436} &
\vstat{0.1260} \\

\midrule

FM & Chronos-2 & \vbase{0.051} &
\vbase{0.048} & \vbase{0.048} &
\vbase{0.047} & \vbase{0.0318} &
\vbase{0.0315} &
\cfgbest\vbaseB{0.0314} &
\cfgbest\vbaseB{0.0319} &
\vbase{0.0577} & \vbase{0.0586} &
\vbase{0.0609} & \vbase{0.0604}
\\

FM & Chronos-2+m & \vmet{0.051} &
\vmet{0.048} & \vmet{0.047} &
\vmet{0.046} & \vmet{0.0341} &
\vmet{0.0348} & \vmetU{0.0343} &
\vmetU{0.0351} & \vmet{0.0599} &
\vmet{0.0605} & \vmet{0.0628} &
\vmet{0.0626} \\

FM & Chronos-2+s+m & \vall{0.053} &
\vall{0.050} & \vall{0.049} &
\vall{0.049} & \vall{0.0387} &
\vall{0.0390} & \vall{0.0383} &
\vall{0.0392} & \vall{0.0641} &
\vall{0.0638} & \vall{0.0662} &
\vall{0.0656} \\

FM & Chronos-2+s & \vstat{0.055} &
\vstat{0.052} & \vstat{0.051} &
\vstat{0.050} & \vstat{0.0411} &
\vstat{0.0408} & \vstat{0.0400} &
\vstat{0.0408} & \vstat{0.0640} &
\vstat{0.0648} & \vstat{0.0671} &
\vstat{0.0663} \\

\bottomrule

\end{tabular}%

}

\end{table}
\vskip -8pt
\subsubsection{UHI Forecasting (Task~2c)}

Table~\ref{tab:1d} shows that \textbf{AirT-UHI remains substantially easier
to predict than LST-UHI.} Complete per-horizon results, including Munich
and Berlin and omitted baselines, are reported in Appendix
Table~\ref{tab:1d-fm-complete}.

Across both UHI sources, XGBoost with meteorological and static features
provides the most robust overall forecasting performance, suggesting
that effective UHI prediction benefits from combining temporal history,
evolving atmospheric conditions, and persistent urban-form information.
However, the optimal model remains city- and source-dependent: in Munich
and Berlin, DLinear variants achieve the lowest LST-UHI MAE, indicating
that regular diurnal and seasonal structure can dominate in some Cfb
cities. This is consistent with the overall results, as the relative
importance of nonlinear meteorological forcing and surface heterogeneity
varies across cities. Covariate benefits are also more consistent in
forecasting than in imputation. Meteorological inputs generally
outperform static features alone, while static morphology provides
complementary information when combined with meteorology. By contrast,
DLinear often degrades with added covariates, showing that auxiliary
inputs help only when the model can effectively exploit heterogeneous
exogenous information.


\begin{table}[t]

\caption{Task~2c forecasting benchmark with average MAE across prediction horizons for Lagos, Johannesburg, Cologne, and Riyadh.}\label{tab:1d}

\small

\setlength{\tabcolsep}{1pt}

\resizebox{\columnwidth}{!}{%

\begin{tabular}{l l c c c c c c c c}

\toprule

& & \multicolumn{4}{c}{Avg MAE AirT-UHI} &
\multicolumn{4}{c}{Avg MAE LST-UHI}
\\

\cmidrule(lr){3-6}\cmidrule(lr){7-10}

Group & Method & Lagos & Johannesburg & Cologne & Riyadh & Lagos
& Johannesburg & Cologne & Riyadh \\

\midrule

ML & XGBoost & \vbase{0.072} &
\vbase{0.135} & \vbase{0.144} &
\vbase{0.097} & \vbase{0.789} &
\vbase{0.836} & \vbase{0.638} &
\vbase{0.808} \\

ML & XGBoost+m & \vmet{0.064} &
\vmetU{0.086} & \vmetU{0.104} &
\vmetU{0.063} & \vmetU{0.741} &
\vmetU{0.784} & \vmetU{0.615} &
\vmetU{0.772} \\

ML & XGBoost+s & \vstat{0.073} &
\vstat{0.123} & \vstat{0.144} &
\vstat{0.097} & \vstat{0.789} &
\vstat{0.834} & \vstat{0.638} &
\vstat{0.808} \\

ML & XGBoost+s+m & \vallU{0.064} &
\cfgbest\vallB{0.083} &
\cfgbest\vallB{0.102} &
\cfgbest\vallB{0.061} &
\cfgbest\vallB{0.737} &
\cfgbest\vallB{0.767} &
\cfgbest\vallB{0.610} &
\cfgbest\vallB{0.763}
\\

\midrule

DL & LSTM & \vbase{0.099} &
\vbase{0.193} & \vbase{0.189} &
\vbase{0.158} & \vbase{1.012} &
\vbase{1.118} & \vbase{0.684} &
\vbase{1.276} \\

DL & DeepUHI & \vbase{0.091} &
\vbase{0.146} & \vbase{0.184} &
\vbase{0.110} & \vbase{0.949} &
\vbase{0.992} & \vbase{0.664} &
\vbase{1.013} \\

\midrule

FM & TimesFM & \vbase{0.068} &
\vbase{0.154} & \vbase{0.200} &
\vbase{0.105} & \vbase{1.266} &
\vbase{0.954} & \vbase{0.662} &
\vbase{0.882} \\

FM & Chronos-2 & \vbase{0.067} &
\vbase{0.159} & \vbase{0.203} &
\vbase{0.105} & \vbase{1.263} &
\vbase{0.923} & \vbase{0.653} &
\vbase{0.853} \\

FM & Chronos-2+m &
\cfgbest\vmetB{0.064} &
\vmet{0.137} & \vmet{0.163} &
\vmet{0.088} & \vmet{1.256} &
\vmet{0.886} & \vmet{0.644} &
\vmet{0.844} \\

\bottomrule

\end{tabular}%

}

\end{table}

\begin{figure}[t]
  \centering
  \includegraphics[width=0.95\linewidth]{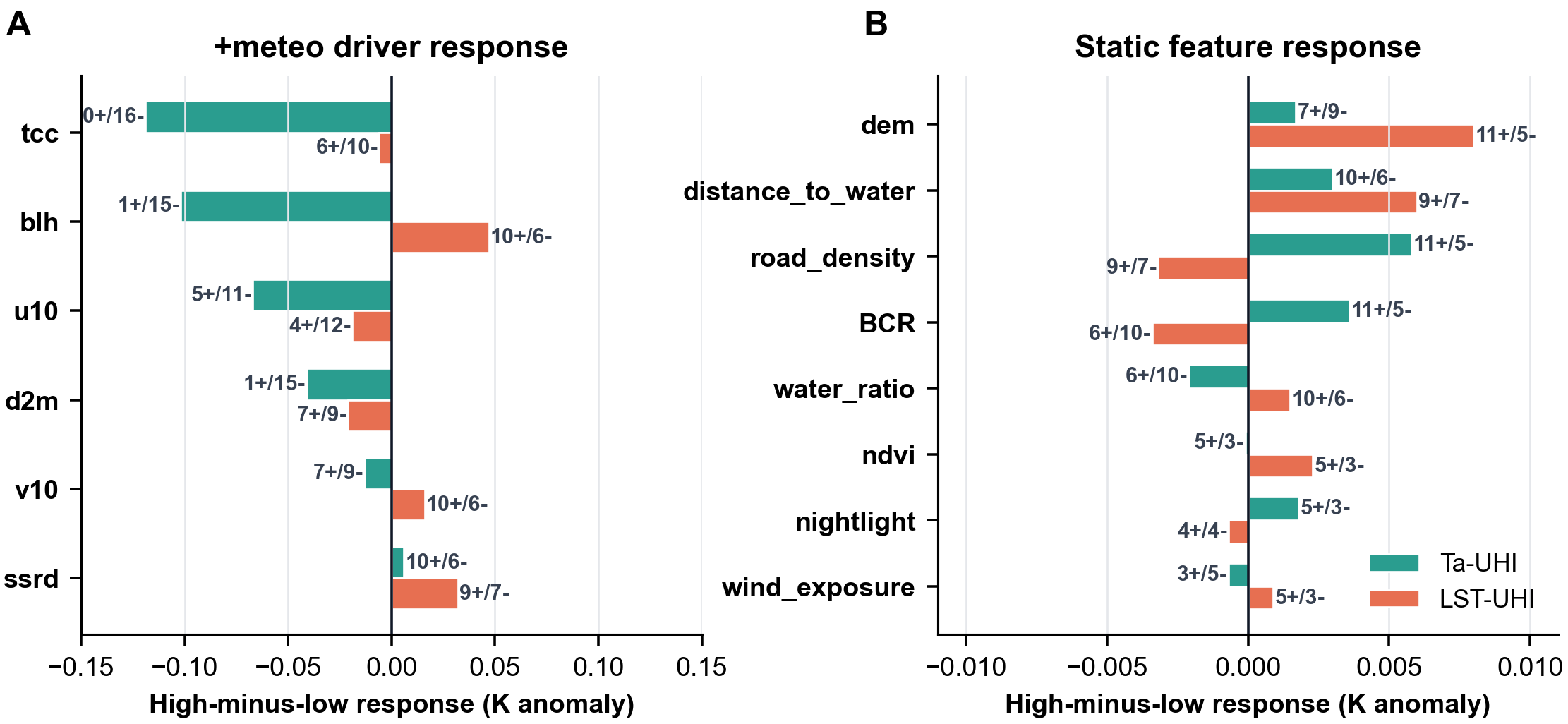}
  \caption{Task~2d Mechanism Stability diagnostics.}
  \label{fig:task2d-mechanism-summary}
  \Description{Two-panel Task 2d mechanism-stability summary with +m driver
  responses and static-feature responses for AirT-UHI and LST-UHI.}
\end{figure}

\subsubsection{Driver Attribution and Mechanism Stability (Task~2d)}

We first assess mechanism stability using eight model
families: Linear Regression, Ridge~\cite{ref76}, Lasso~\cite{ref77},
ElasticNet~\cite{ref78}, Random Forest (RF)~\cite{ref66},
ExtraTrees~\cite{ref79}, HistGradientBoosting~\cite{ref80}, and
XGBoost~\cite{ref65}; detailed Kendall's~$W$
results are reported in Appendix Table~\ref{tab:app2-kw}. These results
show that tree ensembles produce more consistent driver rankings than
linear models.
Their all-hours Kendall's W ranges from 0.626 to 0.659, compared with
0.349--0.515 for linear models. Nighttime rankings are especially
stable, with XGBoost reaching the highest nighttime consistency (W=0.806). We then use XGBoost as the primary method for detailed mechanism diagnostics in Figure~\ref{fig:task2d-mechanism-summary}: Panel~A shows+m driver responses, Panel~B shows static-feature responses, and sign labels count cities with positive/negative responses. Detailed
per-city signed top-3 XGBoost meteorological drivers and top-3 static
features for both targets across all 16 cities are provided in Appendix
Figure~\ref{fig:task2d-city-top3-bars-combined}.

Consistent with established urban-climate
understanding, AirT-UHI exhibits a stable boundary-layer--cloud--wind
structure across the 16 cities, dominated by boundary-layer height,
cloud cover, wind, and humidity. In contrast, LST-UHI has more dispersed
and less stable drivers, and is more strongly associated with static
morphology and surface properties. However, the full Task~2 results show
that driver's importance does not directly determine which auxiliary
inputs improve prediction. Although AirT-UHI is more meteorologically
controlled and LST-UHI are more surface-driven, and meteorological inputs do not
always benefit AirT-UHI more than static features, nor is the reverse
universally true for LST-UHI. Covariate utility is task-dependent:
meteorology improves extreme-event detection for both sources and is
particularly useful for LST-UHI cloud-gap imputation; static
morphology contributes more to AirT-UHI sparse imputation; and
conventional ML models forecast best when both feature groups are
combined.


\subsection{Cross-city and Cross-climate
Transfer (RQ3)}\label{cross-city-and-cross-climate-transfer}

\subsubsection{Climate-diverse source-set transfer (Task~3a).}

For compactness, Table~\ref{tab:app3-full} omits the Stat/Geo, DLinear, and input-sweep rows;
their complete results are reported in Appendix
Table~\ref{tab:app3-full-sweep}.
Table~\ref{tab:app3-full} evaluates whether zero-shot transfer benefits
more from adding additional source cities or from increasing the climate
diversity of the source set. The results show that climate diversity is
more important than simply increasing the number of source cities within
the same climate zone.

\begin{table}[!ht]
  \caption{Task~3 Climate-diverse source-set transfer. \textbf{Bold} =
  overall best, \underline{underline} = overall 2nd.}
  \label{tab:app3-full}
  \centering
  \small
  \setlength{\tabcolsep}{1pt}
  \begin{tabular}{l l c c c}
    \toprule
    & & \multicolumn{3}{c}{MAE (RMSE)} \\
    \cmidrule(lr){3-5}
    Group & Method / source & +1h & +6h & +24h \\
    \midrule
    ML & XGBoost / Cfb-4 (+m) & \vmet{3.390 (4.66)} & \vmet{4.482 (5.77)} & \vmet{3.879 (5.13)} \\
    ML & XGBoost / Cfb-7 (+m) & \vmet{3.185 (4.42)} & \vmet{4.050 (5.28)} & \vmet{3.461 (4.65)} \\
    ML & XGBoost / Diverse-4 & \vbase{0.813 (1.23)} & \vbase{1.174 (1.68)} & \vbaseU{0.983 (1.44)} \\
    ML & XGBoost / Diverse-4 (+m) & \vmet{0.810 (1.22)} & \vmet{1.148 (1.63)} & \vmet{0.988 (1.44)} \\
    ML & XGBoost / Diverse-7 & \vbase{0.797 (1.20)} & \vbase{1.159 (1.65)} & \cfgbest\vbaseB{0.981 (1.43)} \\
    ML & XGBoost / Diverse-7 (+m) & \cfgbest\vmetB{0.794 (1.19)} & \vmetU{1.108 (1.58)} & \vmet{0.988 (1.44)} \\
    \midrule
    FM & Chronos & \vbase{1.000 (1.47)} & \vbase{1.412 (2.00)} & \vbase{1.241 (1.75)} \\
    FM & TimesFM & \vbase{1.017 (1.47)} & \vbase{1.217 (1.74)} & \vbase{1.220 (1.68)} \\
    FM & MOIRAI-1.0-R-small & \vbase{0.947 (1.73)} & \vbase{1.644 (2.66)} & \vbase{2.251 (3.44)} \\
    FM & ChrAdapter / Cfb-4 & \vmet{1.007 (1.41)} & \vmet{2.315 (3.17)} & \vmet{2.116 (2.80)} \\
    FM & ChrAdapter / Cfb-7 & \vmet{1.234 (1.69)} & \vmet{2.001 (2.76)} & \vmet{1.641 (2.27)} \\
    FM & ChrAdapter / Diverse-4 & \vmet{0.796 (1.18)} & \vmet{1.115 (1.58)} & \vmet{0.991 (1.42)} \\
    FM & ChrAdapter / Diverse-7 & \vmetU{0.795 (1.18)} & \cfgbest\vmetB{1.074 (1.52)} & \vmet{1.010 (1.45)} \\
    \bottomrule
  \end{tabular}

  \vspace{2pt}
  \parbox{\columnwidth}{\scriptsize ChrAdapter = Chronos-adapter.}
\end{table}

\begin{figure*}[!t]
  \centering
  \includegraphics[width=\textwidth]{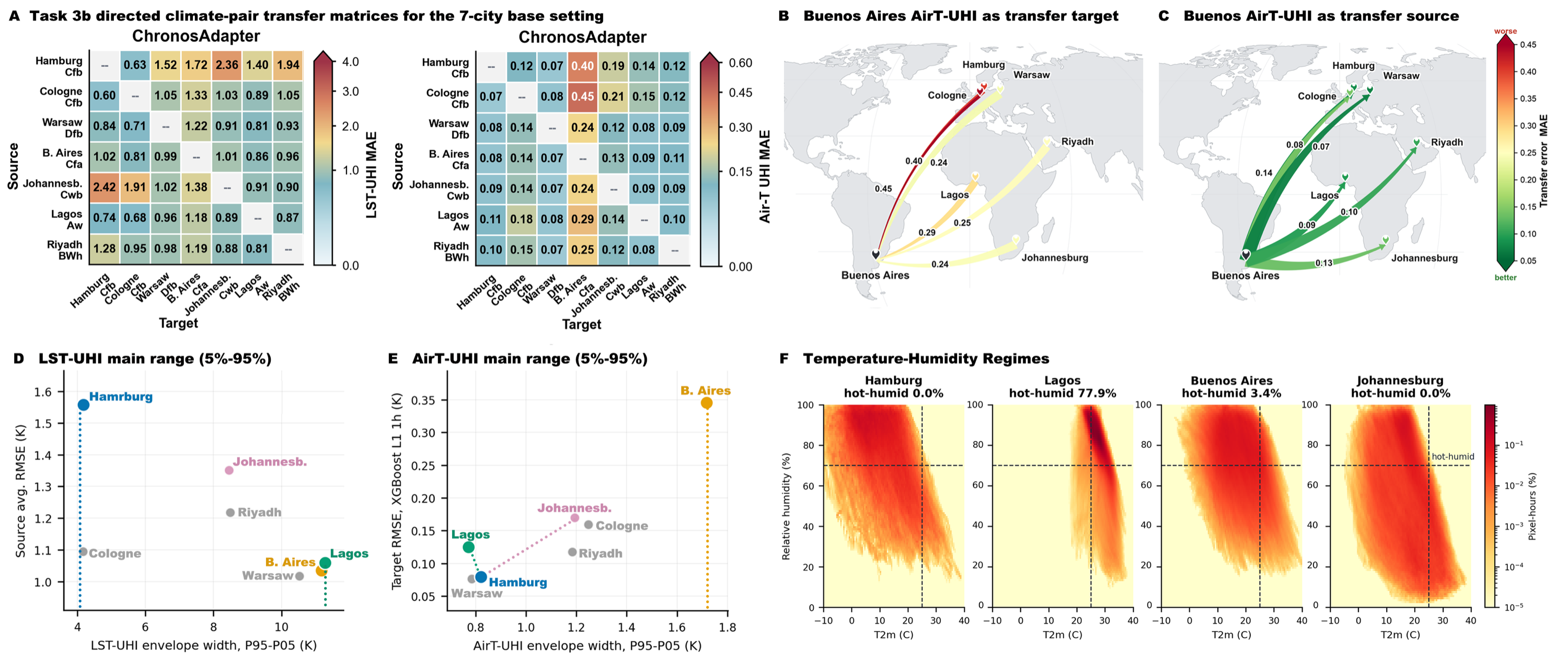}
  \caption{Directed cross-city transfer performance and UHI-regime coverage
  in the seven-city base setting.}
  \label{fig:task3b-climate-pairs}
  \Description{Directed transfer heatmaps for the seven-city base climate
  transfer experiment.}
\end{figure*}

Expanding the homogeneous source set from Cfb-4
to Cfb-7 yields only modest gains, whereas replacing Cfb-4 with
Diverse-4 substantially reduces MAE across XGBoost, ChronosAdapter, and
DLinear. Further expansion from Diverse-4 to Diverse-7 provides little
additional improvement. Thus, transfer performance benefits primarily
from broader climate coverage rather than from adding more source cities
within the same climate regime. The best overall performance is
achieved by climate-diverse source sets. ChronosAdapter with Diverse-7
obtains the lowest average MAE (0.960), closely followed by XGBoost with
Diverse-7 plus static inputs (0.963). These results suggest that
cross-climate representation is critical for zero-shot UHI transfer.
Source cities from multiple climate zones expose the model to a broader
range of meteorological and surface-response regimes, making the
learned patterns more transferable to unseen cities.

\subsubsection{Directed climate-pair transfer (Task~3b)}\label{directed-climate-pair-transfer}

Task 3b evaluates directed zero-shot transfer between individual source--target city pairs. Overall, \textbf{transferability is better explained by overlap in UHI regimes than by Köppen climate-label distance~\cite{ref53}: UHI amplitude, temporal variability, and diurnal behavior matter more than
climate classification alone.} Figure~\ref{fig:task3b-climate-pairs} reports the results under the base ChronosAdapter setting. Appendix Figure~\ref{fig:task3d-input-matrices}
presents the complete directed climate-pair transfer results across all
input configurations, while Appendix Figure~\ref{fig:task3d-transfer-diagnostics}
provides the corresponding transfer-map diagnostics.

For AirT-UHI, transfer is relatively stable across city pairs, and
XGBoost with UHI history alone is the most robust baseline,
since AirT-UHI is smoother and boundary-layer-driven.
Meteorological
and static covariates help transfer directions, but provide no
consistent overall gain, although \texttt{+s+m} helps difficult targets
such as Buenos Aires. As shown in Figure~\ref{fig:task3b-climate-pairs}B,
Buenos Aires is the hardest AirT-UHI target: its average incoming
transfer error is about 105\% higher than that of the next-hardest
target. Figure~\ref{fig:task3b-climate-pairs}E shows that Buenos Aires
also has the widest AirT-UHI main range, consistent with its coastal,
moisture-influenced climate (Figure~\ref{fig:task3b-climate-pairs}F),
its complex, spatially heterogeneous building layout
(Figure~\ref{fig:task3d-transfer-diagnostics}I), and its relatively
high wind speed and moderate boundary-layer height
(Figure~\ref{fig:task3d-transfer-diagnostics}G,~H).
Yet transfer is directed rather than symmetric: Buenos Aires is a
comparatively effective source, as shown in
Figure~\ref{fig:task3b-climate-pairs}C.
Notably, Figure~\ref{fig:task3d-transfer-diagnostics}C shows that Hamburg transfers to Lagos with 26\% lower error than to Johannesburg, despite the larger Köppen contrast: Hamburg and
Lagos have more similar AirT-UHI envelopes
(Figure~\ref{fig:task3b-climate-pairs}E) and seasonal-cycle patterns
(Figure~\ref{fig:task3d-transfer-diagnostics}E) than Hamburg and
Johannesburg.

In contrast, LST-UHI transfer is more sensitive to source--target
mismatch because surface temperature is governed by local surface-energy
balance. ChronosAdapter consistently outperforms conventional
baselines, since LST-UHI depends more on
city-specific surface-energy properties. In Figure~\ref{fig:task3b-climate-pairs}D and
Figure~\ref{fig:task3d-transfer-diagnostics}D, Lagos is the strongest
LST-UHI source, with an average outgoing transfer error about 44\% lower
than Hamburg, because it spans broad surface-heating conditions
(Figure~\ref{fig:task3b-climate-pairs}D) and has a persistently
hot-humid regime (Figure~\ref{fig:task3b-climate-pairs}F).
Hamburg is the weakest source (Figure~\ref{fig:task3d-transfer-diagnostics}A),
with a near-zero LST-UHI distribution
(Figure~\ref{fig:task3d-transfer-diagnostics}F) that poorly supports
targets whose mean LST-UHI is much higher.
Cologne transfers better
than same-Cfb Hamburg, reducing average LST transfer error by about
38\%, mainly because its LST-UHI varies more strongly. Overall, source
selection should prioritize UHI amplitude, temporal variability, and
diurnal-regime coverage over Köppen labels alone.

\vspace{-5pt}

\section{Conclusion}\label{sec:conclusion}\label{conclusion}

In this work, we introduced UHI-Bench, a unified benchmark that integrates LST-UHI, AirT-UHI, hourly meteorological drivers, and static urban morphology features. To the best of our knowledge, UHI-Bench is the first UHI benchmark that supports dual-source UHI modeling. Following a signal, mechanism, and transfer framework, UHI-Bench evaluates over 20 baselines from four model families on five tasks across 20 cities and nine Köppen climate classes. Extensive experiments show that foundation models remain competitive and stable across tasks, motivating the development of domain-specific foundation models for urban heat islands. Environmental covariate utility varies across sources and tasks, highlighting the need for task-aware integration of dynamic meteorology and static urban morphology. Cross-city
transferability is better explained by overlap in UHI regimes than by climate-zone similarity, suggesting that source cities should be selected according to their coverage of target UHI characteristics.                                 
%


\bibliographystyle{ACM-Reference-Format}
\bibliography{references}

\appendix
\clearpage
\onecolumn
\section*{Appendix}

\section{Task 3b Transfer Diagnostics}
\label{sec:task3b-transfer-diagnostics}

\begin{center}
  \includegraphics[width=\textwidth,height=0.88\textheight,keepaspectratio]{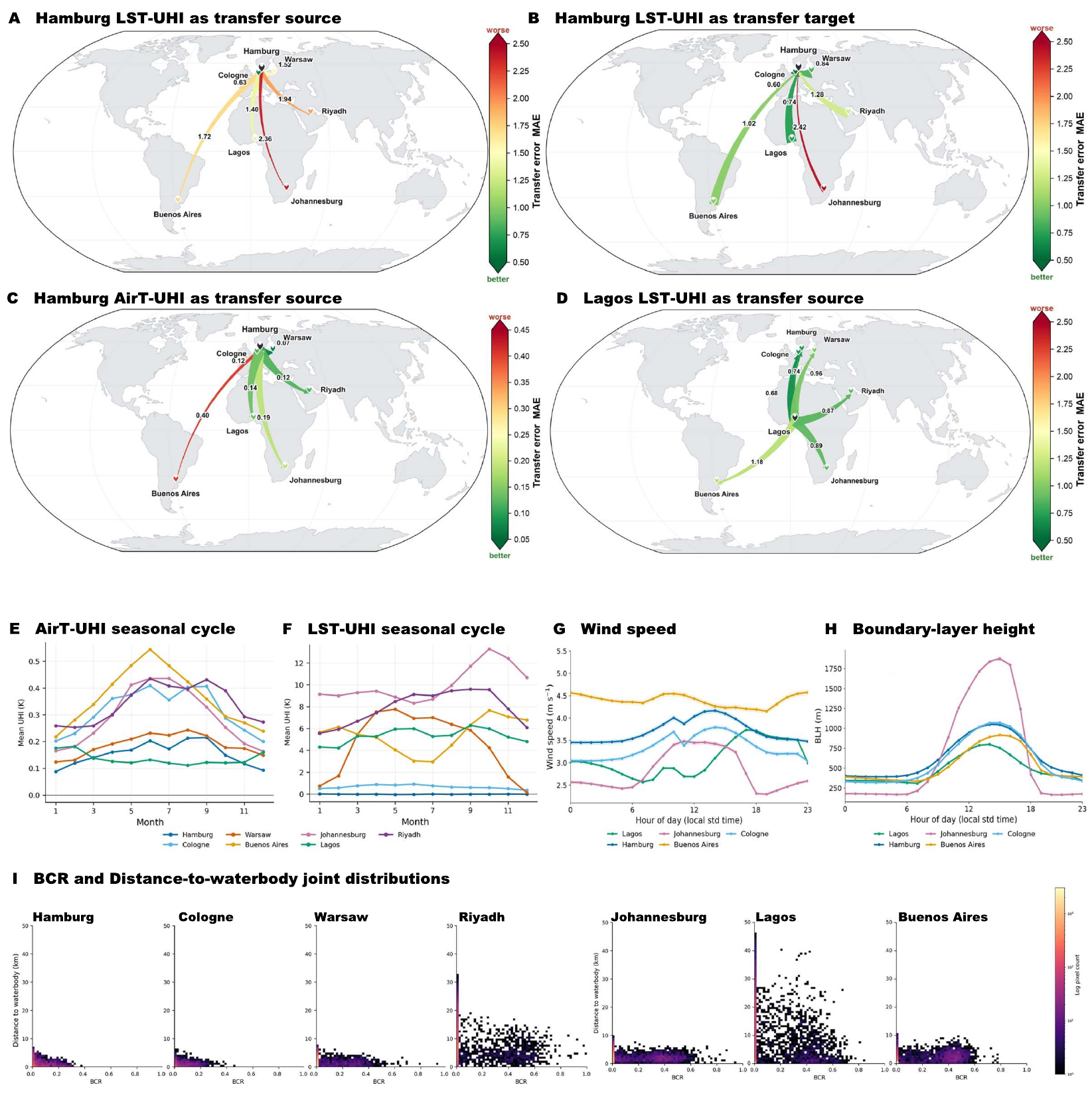}
  \captionof{figure}{Task~3b transfer-map diagnostics and UHI-regime summaries.}
  \label{fig:task3d-transfer-diagnostics}
\end{center}

\clearpage

\begin{center}
  \includegraphics[width=\textwidth,height=0.88\textheight,keepaspectratio]{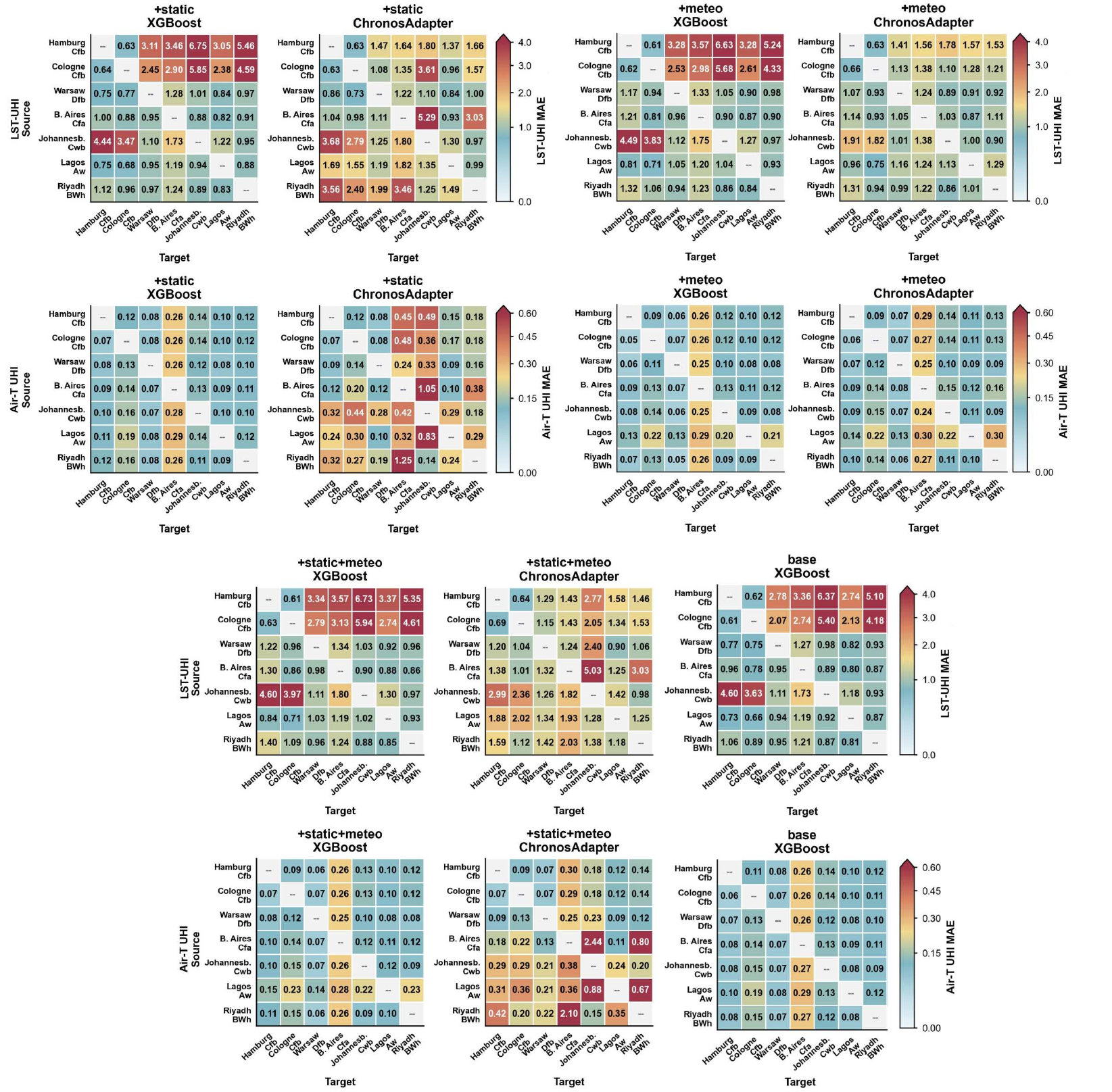}
  \captionof{figure}{Task~3b directed climate-pair transfer complete results}
  \label{fig:task3d-input-matrices}
\end{center}

\clearpage

\section{Full Dataset Summary}
\label{sec:dataset-summary}

{\scriptsize
\setlength{\LTcapwidth}{\textwidth}
\setlength{\tabcolsep}{1.4pt}
\renewcommand{\arraystretch}{0.98}
\input{appendix_dataset_summary_from_source.tex}
}

\begin{figure}[!htbp]
  \centering
  \includegraphics[width=0.90\textwidth]{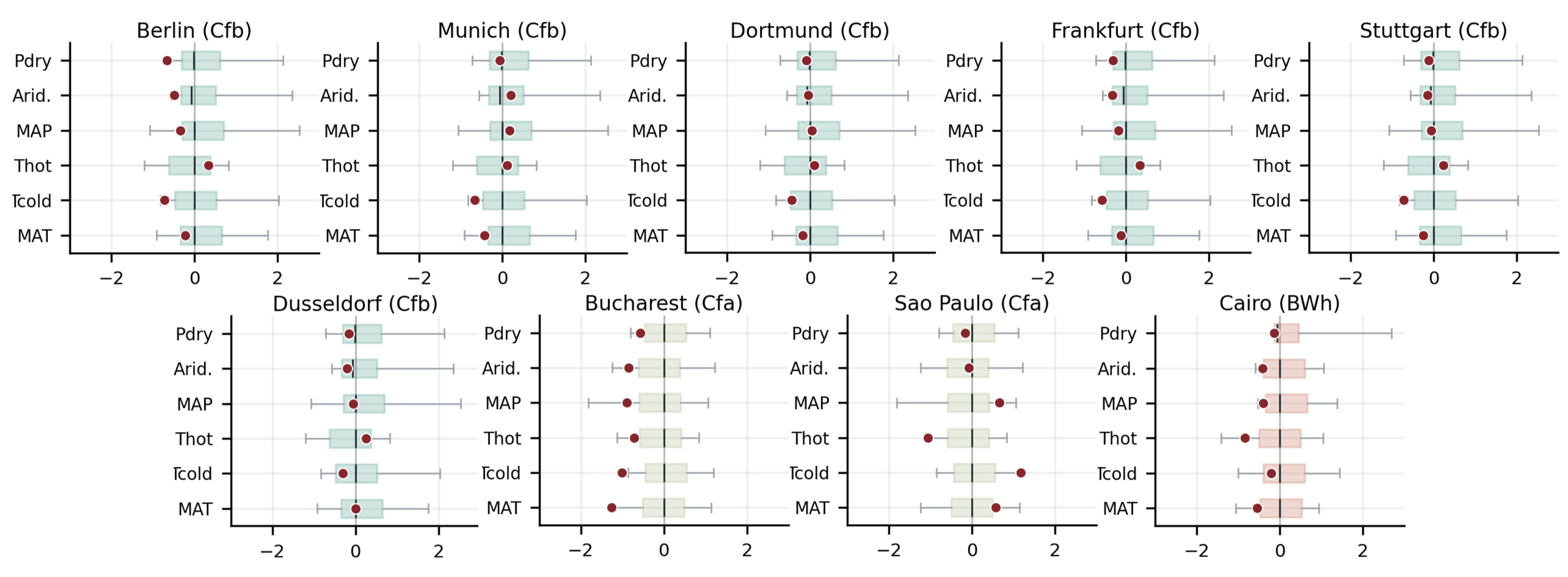}
  \caption{Relative distributions of K\"oppen--Geiger climate classification indicators
  for additional representative cities. The red markers show each city, and the
  shaded distributions provide benchmark-wide context.}
  \label{fig:climate-coverage-appendix}
  \Description{Appendix continuation of the climate-coverage figure, showing
  relative distributions of climate classification indicators for Berlin,
  Munich, Dortmund, Dusseldorf, Frankfurt, Stuttgart, Bucharest, Sao Paulo, and
  Cairo.}
\end{figure}

\clearpage
\twocolumn

\section{Related UHI Modeling Methods}
\label{sec:app-uhi-modeling}

Early UHI studies relied mainly on correlation, regression, and
geostatistical methods to quantify urban--rural temperature
differences and interpolate sparse
observations~\cite{ref23}. Methods such as inverse-distance weighting and kriging
remain effective under strong spatial autocorrelation, but their simple
assumptions limit nonlinear and long-term dependencies. Machine-learning
models, especially RF~\cite{ref66} and XGBoost~\cite{ref65}, were later adopted for UHI
prediction, imputation, and driver analysis. They integrate
heterogeneous features and support
interpretation via feature importance and SHAP. However, most
studies evaluate them for a single city, temperature source, or task,
limiting cross-study and cross-climate comparison.

More recently, UHI modeling has been formulated as multivariate
\mbox{time-series} or spatiotemporal forecasting. Models such as DLinear~\cite{ref70},
Autoformer\footnote{\scriptsize Haixu Wu, Jiehui Xu, Jianmin Wang, and Mingsheng Long. 2021. Autoformer: Decomposition Transformers with Auto-Correlation for Long-Term Series Forecasting. NeurIPS 2021.}, PatchTST~\cite{ref69}, iTransformer\footnote{\scriptsize Yong Liu, Tengge Hu, Haoran Zhang, Haixu Wu, Shiyu Wang, Lintao Ma, and Mingsheng Long. 2024. iTransformer: Inverted Transformers Are Effective for Time Series Forecasting. ICLR 2024.},
STGCN\footnote{\scriptsize Bing Yu, Haoteng Yin, and Zhanxing Zhu. 2018. Spatio-Temporal Graph Convolutional Networks: A Deep Learning Framework for Traffic Forecasting. IJCAI 2018, 3634--3640.}, and STID\footnote{\scriptsize Zezhi Shao, Zhao Zhang, Fei Wang, Wei Wei, and Yongjun Xu. 2022. Spatial-Temporal Identity: A Simple yet Effective Baseline for Multivariate Time Series Forecasting. CIKM 2022, 4454--4458.} capture temporal
patterns and spatial dependencies for forecasting and gap filling.
Context-aware methods such as DeepUHI~\cite{ref33} further support fine-grained
station-temperature prediction. Yet these are usually tied
to specific cities or target definitions. Earth-
observation models such as Prithvi-EO\footnote{\scriptsize Johannes Jakubik, Sujit Roy, et al. 2023. Foundation Models for Generalist Geospatial Artificial Intelligence. NeurIPS 2023 Workshop on Foundation Models for Decision Making.}, Granite-LST\footnote{\scriptsize Jannis Fleckenstein, David Kreismann, Tamara Rosemary Govindasamy, Thomas Brunschwiler, Etienne Vos, and Mattia Rigotti. 2025. Detection and Simulation of Urban Heat Islands Using a Fine-Tuned Geospatial Foundation Model for Microclimate Impact Prediction. NeurIPS 2025 Workshop on Tackling Climate Change with Machine Learning.},
and Prithvi-WxC\footnote{\scriptsize Johannes Schmude, Sujit Roy, et al. 2024. Prithvi WxC: A Weather and Climate Foundation Model. AGU 2024.}
operate on imagery, absolute LST, or atmospheric fields but do not
produce UHI anomalies without urban--rural referencing. General
time-series foundation models such as Chronos~\cite{ref62},
TimesFM~\cite{ref63}, and MOIRAI~\cite{ref64} can
be evaluated directly on UHI sequences, but their suitability for
different UHI sources and tasks remains unclear. Mechanism analysis and
cross-city generalization remain fragmented, rarely comparing stable
drivers across LST-UHI and AirT-UHI or evaluating directed transfer
under different climate and input settings. UHI-Bench
addresses these gaps by jointly benchmarking statistical,
geostatistical, machine-learning, deep-learning, and time-series
foundation models. It evaluates which models suit different UHI tasks,
when meteorological and static features are useful, how the two
UHI sources differ, and what determines successful transfer to target
cities.

\section{Data Processing}
\label{sec:data-processing}

\subsection{UHI Target Variable Computation}
\label{sec:uhi-computation}

This appendix formalizes the LST-UHI and AirT-UHI construction
procedures described in
Section~\ref{data-sources-and-construction}, indexing pixel $p$ and
timestamp $t$.

\subsubsection{LST-UHI}\label{sec:app-lst-uhi}

The native MSG/SEVIRI LST product (nominal 3~km resolution) is
downscaled to 1~km using an RF-TsHARP (random-forest thermal sharpening)
model:
\[
\text{LST}^{1\text{km}}_{p,t} = f_{\text{RF-TsHARP}}\!\left(
\text{LST}^{3\text{km}}_{p,t},\, X_{\text{morph},p}\right),
\]
where $X_{\text{morph},p}$ denotes the auxiliary morphological/spectral
predictors (e.g., built-up coverage, NDVI, water bodies) used to sharpen
the coarse-resolution LST at pixel $p$. Downscaling quality is assessed
via 3~km round-trip consistency and comparison with MODIS 1~km LST.

LST-UHI is then defined as the local downscaled LST minus the
contemporaneous mean LST of surrounding rural reference pixels:
\[
\text{LST-UHI}_{p,t} = \text{LST}^{1\text{km}}_{p,t} -
\overline{\text{LST}}_{\mathcal{R}_c,t}, \qquad
\overline{\text{LST}}_{\mathcal{R}_c,t} = \frac{1}{|\mathcal{R}_c|}
\sum_{r \in \mathcal{R}_c} \text{LST}^{1\text{km}}_{r,t},
\]
where $\mathcal{R}_c$ is the rural reference set for the city $c$ that
pixel $p$ belongs to (defined below).
Cloud-contaminated observations are retained as missing values and are
not filled.

\paragraph{Rural reference pixels}
For each city $c$, rural reference pixels are selected from an annulus
15--25~km away from the city center. Let $(\phi_p,\lambda_p)$ and
$(\phi_c,\lambda_c)$ denote the latitude and longitude of pixel $p$ and
city center $c$. The great-circle distance is computed as
\[
d_{p,c}
=
2R_{\oplus}\arcsin
\sqrt{
\sin^2\!\left(\frac{\phi_p-\phi_c}{2}\right)
+
\cos\phi_p\cos\phi_c
\sin^2\!\left(\frac{\lambda_p-\lambda_c}{2}\right)
},
\]
where $R_{\oplus}$ is the Earth radius. The rural reference set is
\[
\mathcal{R}_{c}
=
\left\{
p:
15 \le d_{p,c} \le 25,\;
\mathrm{LC}_{p}\in\{\mathrm{cropland},\mathrm{grassland}\}
\right\},
\]
with distances $d_{p,c}$ in kilometers and $\mathrm{LC}_p$ the land-cover
class of pixel $p$. Built-up and water pixels are excluded from the
rural reference set.\par
\noindent\parbox{\columnwidth}{\scriptsize
\textit{Notes.}
$p$: 1~km grid pixel; $t$: UTC-hour timestamp; $c$: city.
$\text{LST}^{3\text{km}}_{p,t}$ and $\text{LST}^{1\text{km}}_{p,t}$:
native and downscaled MSG/SEVIRI LST. $f_{\text{RF-TsHARP}}(\cdot)$:
trained random-forest thermal-sharpening model; $X_{\text{morph},p}$:
morphological/spectral predictors. $\mathcal{R}_c$: rural reference
set; $\overline{\text{LST}}_{\mathcal{R}_c,t}$: mean LST over
$\mathcal{R}_c$ at time $t$. $(\phi_p,\lambda_p)$ and
$(\phi_c,\lambda_c)$: pixel/city-center coordinates; $d_{p,c}$:
great-circle distance in km; $R_{\oplus}$: Earth radius;
$\mathrm{LC}_p$: land-cover class.}

\subsubsection{AirT-UHI}\label{sec:app-airt-uhi}

For the eight German cities, AirT is taken directly from the DWD
HOSTRADA gridded 2~m air-temperature product:
\[
\text{AirT}_{p,t} = \text{HOSTRADA}_{p,t}.
\]

For the eight international cities, where comparable long-term,
spatially continuous AirT observations are unavailable, AirT is instead
produced by a residual correction model trained on
HOSTRADA$-$ERA5 residuals from the German cities:
\[
\begin{aligned}
\text{AirT}_{p,t}
&= \text{ERA5}_{p,t} \\
&\quad + f_{\theta}\!\left(
\text{ERA5}_{p,t},\, X_{\text{morph},p},\,
X_{\text{temporal},t}\right),
\end{aligned}
\]
where $f_{\theta}$ is trained to predict the HOSTRADA$-$ERA5 residual
from meteorological, morphological, and temporal predictors, following
geographically-weighted-regression-based strategies for merging
coarse-resolution gridded products with ground observations. The model
achieves held-out $R^2 \approx 0.79$.

AirT-UHI is then defined as the local air temperature minus the mean
air temperature over a 15--25~km rural annulus surrounding the city,
which also removes biases common to both urban and rural pixels:
\[
\begin{aligned}
\text{AirT-UHI}_{p,t}
&= \text{AirT}_{p,t} -
\overline{\text{AirT}}_{\mathcal{A}_c,t}, \\
\overline{\text{AirT}}_{\mathcal{A}_c,t}
&= \frac{1}{|\mathcal{A}_c|}
\sum_{a \in \mathcal{A}_c} \text{AirT}_{a,t},
\end{aligned}
\]
where $\mathcal{A}_{c} = \{p : 15 \le d_{p,c} \le 25\}$ is the set of
pixels within the 15--25~km annulus around city center $c$, using the
same great-circle distance $d_{p,c}$ defined above for $\mathcal{R}_c$
(without the cropland/grassland land-cover restriction).

\par
\noindent\parbox{\columnwidth}{\scriptsize
\textit{Notes.}
$p$: a 1~km grid pixel; $t$: a timestamp (UTC hour); $c$: a city.
$\text{HOSTRADA}_{p,t}$: the DWD HOSTRADA gridded 2~m
air-temperature observation at pixel $p$, time $t$ (German cities
only). $\text{ERA5}_{p,t}$: the native ERA5 air-temperature value at
pixel $p$, time $t$; $f_{\theta}(\cdot)$: the trained residual
correction model; $X_{\text{morph},p}$: morphological predictors at
pixel $p$; $X_{\text{temporal},t}$: temporal predictors at time $t$;
$R^2$: held-out coefficient of determination of $f_{\theta}$.
$\mathcal{A}_c$: the 15--25~km rural annulus pixel set for city
$c$ (same great-circle distance $d_{p,c}$ as $\mathcal{R}_c$, without
the land-cover restriction);
$\overline{\text{AirT}}_{\mathcal{A}_c,t}$: mean air temperature over
$\mathcal{A}_c$ at time $t$.}

\subsection{Spatiotemporal Alignment}
\label{sec:spatiotemporal-alignment}

All spatiotemporal co-registration is performed on absolute UTC time and
a common 1 km city grid. Raw hourly timestamps from MSG/SEVIRI LST,
HOSTRADA AirT-UHI, corrected AirT-UHI, and ERA5 are stored as
timezone-naive datetimes but interpreted consistently as UTC; all source
matching, model inputs, train/test splits, and ERA5 joins use exact UTC
hourly timestamps. Spatially, each city is represented by a fixed 1 km
grid with stable pixel\_ids: ERA5 variables are bilinearly interpolated
to the 1 km pixel centres, MSG LST is downscaled or resampled onto the
same grid, and AirT-UHI products are mapped to the same pixel IDs before
any cross-source comparison. This UTC-based alignment avoids ambiguity
from daylight-saving transitions and ensures that ERA5, satellite, and
AirT-UHI records refer to the same physical hour. For the heat-risk
timing diagnostics only, after extreme labels are constructed on the
UTC-aligned series, timestamps are converted from UTC to each city's
local civil time using the corresponding IANA time zone, and
hour-of-day/day-night summaries are reported in local hour. Thus the
timing plots describe local heat-risk phase, while the underlying data
fusion and model evaluation remain UTC-aligned.

\begin{table*}[!t]
  \caption{Aggregation formulas for the static grid features used in
  UHI-Bench.}
  \label{tab:feature-formulas}
  \small
  \setlength{\tabcolsep}{4pt}
  \renewcommand{\arraystretch}{1.4}
  \resizebox{\textwidth}{!}{%
  \begin{tabular}{%
    >{\raggedright\arraybackslash}p{2.6cm}
    >{\raggedright\arraybackslash}p{2.6cm}
    >{\raggedright\arraybackslash}p{9.5cm}
    >{\centering\arraybackslash}p{2.0cm}
  }
    \toprule
    Feature & Aggregation Method & Formula & Output Unit \\
    \midrule
    NDVI &
    Area-weighted mean (NDVI $>$ 0) &
    $\displaystyle \text{NDVI}_i = \frac{\sum_{p_n \in P_i^{+}}
    \text{NDVI}_{p_n} \cdot A_{p_n \cap i}}{\sum_{p_n \in P_i^{+}}
    A_{p_n \cap i}}$, where $P_i^{+} = \{p_n \mid \text{NDVI}_{p_n} > 0\}$.
    If $P_i^{+} = \emptyset$, $\text{NDVI}_i = 0$. &
    -- \\
    \addlinespace
    BCR &
    Area ratio &
    $\displaystyle \text{BCR}_i = \frac{\sum_{j=1}^{M} A_{b_j \cap i}}
    {A_{\text{grid},i}}$ &
    -- \\
    \addlinespace
    Building Height &
    Area-weighted mean &
    $\displaystyle H_i = \frac{\sum_{j=1}^{M} h_j \cdot A_{b_j \cap i}}
    {\sum_{j=1}^{M} A_{b_j \cap i}}$ &
    m \\
    \addlinespace
    Water Surface Ratio &
    Area ratio &
    $\displaystyle W_i = \frac{\sum_{j=1}^{M} A_{w_j \cap i}}
    {A_{\text{grid},i}}$ &
    -- \\
    \addlinespace
    Waterbody Proximity &
    Minimum distance &
    $\displaystyle D_{\text{water},i} = \min_j \{d(P_i, W_j)\}$ &
    m \\
    \addlinespace
    Road Density &
    Length density &
    $\displaystyle \text{RD}_i = \frac{\sum_{j=1}^{M} L_{r_j \cap i}}
    {A_{\text{grid},i}} \times 1000$ &
    km/km$^2$ \\
    \addlinespace
    POI Density &
    Point density &
    $\displaystyle \text{POI}_i = \frac{N_{\text{poi},i}}
    {A_{\text{grid},i}} \times 10^6$ &
    points/km$^2$ \\
    \addlinespace
    Nighttime Light &
    Temporal mean + pixel alignment &
    $\displaystyle \text{NL}_{p,y} = A_p\!\left[|M_y|^{-1}
    \sum_{m \in M_y} \text{VIIRS}^{avg\_rad}_m\right]$ &
    nW/cm$^2$/sr \\
    \addlinespace
    Elevation (DEM) &
    Pixel alignment &
    $\displaystyle \text{DEM}_p = A_p\!\left[\text{DEM}_{GLO30}\right]$ &
    m \\
    \addlinespace
    Wind Exposure &
    Derived proxy &
    $\displaystyle \text{WE}_{p,y} = s_{p,y} \left(1 - \text{BCR}^{*}_p
    \right)$, where $s_{p,y} = \sqrt{\bar{u}_{10,p,y}^2 +
    \bar{v}_{10,p,y}^2}$ and $\text{BCR}^{*}_p = \dfrac{\text{BCR}_p -
    \min_c \text{BCR}}{\max_c \text{BCR} - \min_c \text{BCR} +
    10^{-6}}$ &
    m/s \\
    \bottomrule
	  \end{tabular}%
	  }
\vspace{2pt}
\noindent\parbox{\textwidth}{\scriptsize
\textit{Notes.}
$i$: standardized 1~km reporting grid cell; $p_n$: source raster pixel
(e.g., NDVI); $b_j$: building footprint polygon; $w_j$: waterbody
polygon; $r_j$: road segment. $A_{x \cap i}$: intersection area between
geometry $x$ and grid cell $i$; $A_{\text{grid},i}$: grid-cell area.
$P_i$: grid-cell centroid; $h_j$: building height; $d(\cdot,\cdot)$:
Euclidean distance; $N_{\text{poi},i}$: count of POIs intersecting grid
cell $i$. $p$: 1~km grid pixel for Nighttime Light, Elevation, and Wind
Exposure; $A_p[\cdot]$: raster-to-pixel resampling/alignment operator.
$M_y$: calendar months in year $y$; $\text{VIIRS}^{avg\_rad}_m$: VIIRS
monthly average-radiance composite; $\text{DEM}_{GLO30}$: Copernicus
GLO-30 DEM. $s_{p,y}$: annual-mean 10~m wind-speed magnitude from
ERA5-Land $\bar{u}_{10,p,y}$ and $\bar{v}_{10,p,y}$;
$\text{BCR}^{*}_p$: city-wise min--max-normalized BCR, so higher
building coverage discounts raw wind-speed exposure.}
\end{table*}

\begin{table*}[!t]
  \caption{Static urban morphology features used in UHI-Bench.}
  \label{tab:glossary-static}
  \fontsize{7.5pt}{8.5pt}\selectfont
  \renewcommand{\arraystretch}{0.92}
  \begin{tabular}{p{3cm} p{10.3cm}}
    \toprule
    Feature & Definition \\
    \midrule
    Building coverage ratio (BCR) &
    the proportion of grid area covered by building footprints. \\
    \addlinespace
    Mean building height &
    the area-weighted average height of all buildings intersecting a
    grid cell, giving greater influence to larger structures. \\
    \addlinespace
    Normalized Difference Vegetation Index (NDVI) &
    seasonal mean greenness calculated from Sentinel-2 imagery and
    aggregated using area-weighted averaging across all intersecting
    pixels. MODIS LST is used separately for cross-validation of the
    downscaled LST product (Section~\ref{sec:bench}) and is not the
    source of the NDVI feature. \\
    \addlinespace
    Water coverage ratio &
    the proportion of grid area covered by water surface polygons,
    calculated as the intersection area between waterbody polygons and
    the grid cell divided by the total grid area. \\
    \addlinespace
    Distance to nearest waterbody &
    Euclidean distance from the grid centroid to the nearest lake/pond
    (and coastal) water polygon. \\
    \addlinespace
    POI density &
    number of Points of Interest per square kilometer, used as a proxy
    for human activity intensity and land-use mix. \\
    \addlinespace
    Road density &
    total length of roads within each grid normalized by its area
    (kilometers of road per km$^2$). \\
    \addlinespace
    Nighttime light &
    VIIRS-derived night-time radiance intensity within each grid cell,
    used as a proxy for human activity intensity and
    built-up/economic development. \\
    \addlinespace
    Elevation &
    grid-cell terrain elevation derived from a digital elevation model
    (DEM). \\
    \addlinespace
    Wind exposure (proxy) &
    a grid-level proxy for ventilation potential derived from
    ERA5-Land wind speed, used to characterize local airflow
    conditions relevant to heat dissipation. \\
    \bottomrule
\end{tabular}
\end{table*}

\begin{table*}[!t]
  \caption{Meteorological drivers (ERA5-Land) used in UHI-Bench. All
  six variables are obtained from ERA5-Land and resampled from their
  native grid to the standardized 1~km city grid
  (Section~\ref{data-sources-and-construction}).}
  \label{tab:glossary-meteo}
  \small
  \renewcommand{\arraystretch}{1.0}
  \begin{tabular}{p{3cm} p{10.3cm}}
    \toprule
    Variable & Definition \\
    \midrule
    $u10$, $v10$ (10 m zonal and meridional wind) &
    the eastward and northward components of wind velocity at 10 m
    above the surface, respectively. \\
    \addlinespace
    $tcc$ (total cloud cover) &
    the fraction of the grid cell covered by cloud, ranging from 0
    (clear sky) to 1 (fully overcast). \\
    \addlinespace
    $d2m$ (2 m dewpoint temperature) &
    the temperature to which air at 2 m above the surface would need
    to be cooled to reach saturation, used as a proxy for
    near-surface humidity. \\
    \addlinespace
    $blh$ (boundary-layer height) &
    the height of the atmospheric boundary layer, reflecting the
    degree of vertical mixing near the surface. \\
    \addlinespace
    $ssrd$ (surface solar radiation downwards) &
    the cumulative downward shortwave radiation reaching the
    surface. \\
    \bottomrule
  \end{tabular}

\vspace{4pt}
\noindent
\begin{minipage}[t]{0.45\textwidth}
  \captionof{table}{Task~2d model-family consistency.}
  \label{tab:app2-kw}
  \scriptsize
  \setlength{\tabcolsep}{3pt}
  \centering
  \resizebox{\linewidth}{!}{%
  \begin{tabular}{l c c c}
    \toprule
    Model & all-hours & daytime & nighttime \\
    \midrule
    Linear Regression & 0.422 & 0.426 & 0.605 \\
    Ridge & 0.349 & 0.411 & 0.576 \\
    Lasso & 0.515 & 0.367 & 0.476 \\
    ElasticNet & 0.471 & 0.376 & 0.402 \\
    RF & 0.642 & 0.568 & 0.724 \\
    ExtraTrees & 0.643 & 0.588 & 0.739 \\
    HistGradientBoosting & 0.626 & 0.612 & 0.524 \\
    \textbf{XGBoost} & \textbf{0.659} & 0.475 & \textbf{0.806} \\
    \bottomrule
  \end{tabular}%
  }
  \par\vspace{1pt}
  \scriptsize\noindent\textit{Notes.} Kendall~$W$ of the per-city
  driver ranking across all 16 cities, by model and day/night.
\end{minipage}\hfill
\begin{minipage}[t]{0.5\textwidth}
  \captionof{table}{Complete Task~3a climate-diverse source-set transfer.}
  \label{tab:app3-full-sweep}

  \tiny
  \renewcommand{\arraystretch}{0.82}
  \setlength{\tabcolsep}{0.5pt}
  \centering
  \resizebox{\linewidth}{!}{%
  \begin{tabular}{l l c c c}
    \toprule
    & & \multicolumn{3}{c}{MAE (RMSE)} \\
    \cmidrule(lr){3-5}
    Group & Method / source & +1h & +6h & +24h \\
    \midrule
    Stat/Geo & Persistence & \vbase{0.900 (1.40)} & \vbase{1.645 (2.39)} & \vbase{1.244 (1.84)} \\
    Stat/Geo & SrcClim / Cfb-4 & \vbase{6.709 (7.98)} & \vbase{6.693 (7.95)} & \vbase{6.717 (7.99)} \\
    Stat/Geo & SrcClim / Cfb-7 & \vbase{6.650 (7.92)} & \vbase{6.623 (7.88)} & \vbase{6.668 (7.93)} \\
    Stat/Geo & SrcClim / Diverse-4 & \vbase{4.451 (5.43)} & \vbase{4.446 (5.43)} & \vbase{4.441 (5.44)} \\
    Stat/Geo & SrcClim / Diverse-7 & \vbase{4.384 (5.26)} & \vbase{4.359 (5.22)} & \vbase{4.328 (5.23)} \\
    \midrule
    ML & XGBoost / Cfb-4 & \vbase{3.143 (4.36)} & \vbase{4.153 (5.44)} & \vbase{3.591 (4.81)} \\
    ML & XGBoost / Cfb-4 (+m) & \vmet{3.390 (4.66)} & \vmet{4.482 (5.77)} & \vmet{3.879 (5.13)} \\
    ML & XGBoost / Cfb-4 (+s+m) & \vall{3.443 (4.71)} & \vall{4.671 (5.95)} & \vall{4.021 (5.27)} \\
    ML & XGBoost / Cfb-7 & \vbase{3.068 (4.27)} & \vbase{3.967 (5.22)} & \vbase{3.405 (4.66)} \\
    ML & XGBoost / Cfb-7 (+m) & \vmet{3.185 (4.42)} & \vmet{4.050 (5.28)} & \vmet{3.461 (4.65)} \\
    ML & XGBoost / Cfb-7 (+s+m) & \vall{3.233 (4.47)} & \vall{4.367 (5.64)} & \vall{3.844 (5.08)} \\
    ML & XGBoost / Diverse-4 & \vbase{0.813 (1.23)} & \vbase{1.174 (1.68)} & \vbaseU{0.983 (1.44)} \\
    ML & XGBoost / Diverse-4 (+m) & \vmet{0.810 (1.22)} & \vmet{1.148 (1.63)} & \vmet{0.988 (1.44)} \\
    ML & XGBoost / Diverse-4 (+s+m) & \vall{0.811 (1.22)} & \vall{1.148 (1.61)} & \vall{0.992 (1.44)} \\
    ML & XGBoost / Diverse-7 & \vbase{0.797 (1.20)} & \vbase{1.159 (1.65)} & \cfgbest\vbaseB{0.981 (1.43)} \\
    ML & XGBoost / Diverse-7 (+m) & \cfgbest\vmetB{0.794 (1.19)} & \vmet{1.108 (1.58)} & \vmet{0.988 (1.44)} \\
    ML & XGBoost / Diverse-7 (+s+m) & \vall{0.796 (1.20)} & \vall{1.114 (1.58)} & \vall{0.987 (1.44)} \\
    \midrule
    FM & Chronos & \vbase{1.000 (1.47)} & \vbase{1.412 (2.00)} & \vbase{1.241 (1.75)} \\
    FM & TimesFM & \vbase{1.017 (1.47)} & \vbase{1.217 (1.74)} & \vbase{1.220 (1.68)} \\
    FM & MOIRAI-1.0-R-small & \vbase{0.947 (1.73)} & \vbase{1.644 (2.66)} & \vbase{2.251 (3.44)} \\
    FM & ChrAdapter / Cfb-4 & \vmet{1.007 (1.41)} & \vmet{2.315 (3.17)} & \vmet{2.116 (2.80)} \\
    FM & ChrAdapter / Cfb-7 & \vmet{1.234 (1.69)} & \vmet{2.001 (2.76)} & \vmet{1.641 (2.27)} \\
    FM & ChrAdapter / Diverse-4 & \vmet{0.796 (1.18)} & \vmet{1.115 (1.58)} & \vmet{0.991 (1.42)} \\
    FM & ChrAdapter / Diverse-7 & \vmetU{0.795 (1.18)} & \cfgbest\vmetB{1.074 (1.52)} & \vmet{1.010 (1.45)} \\
    FM & ChrAdapter / Diverse-8 & \vmet{0.814 (1.20)} & \vmetU{1.085 (1.54)} & \vmet{0.989 (1.43)} \\
    \midrule
    DL & DLinear / Cfb-4 & \vbase{3.231 (3.68)} & \vbase{3.964 (4.40)} & \vbase{3.728 (4.16)} \\
    DL & DLinear / Cfb-7 & \vbase{5.335 (5.81)} & \vbase{6.066 (6.59)} & \vbase{5.739 (6.24)} \\
    DL & DLinear / Diverse-4 & \vbase{1.350 (1.84)} & \vbase{1.480 (2.00)} & \vbase{1.359 (1.85)} \\
    DL & DLinear / Diverse-7 & \vbase{1.232 (1.67)} & \vbase{1.412 (1.91)} & \vbase{1.386 (1.88)} \\
    \bottomrule
  \end{tabular}%
  }
  \par\vspace{1pt}
  \scriptsize\noindent\textit{Notes.} SrcClim = SourceClimatology;
  ChrAdapter = Chronos-adapter.
\end{minipage}
\end{table*}
\section{Evaluation Setup and Supplementary Results}
\label{sec:eval-supp}

\subsection{Experimental Setup Details}
\label{sec:app-experimental-setup}

This appendix details the evaluation tasks and baselines built on the
UHI-Bench assets introduced in Section~\ref{sec:bench}. Reported
experiments use representative city subsets selected by data
availability, quality, missingness, and climate diversity; the benchmark
can be extended to additional cities and climate zones. All tasks follow
the temporal splits and leakage-control protocol in Section~3.3, with
task settings and metrics summarized in Table~\ref{tab:task-overview}.

UHI-Bench follows a signal--mechanism--transfer logic. Analysis~1a,
Analysis~1b, and Task~1 first test whether LST-UHI and AirT-UHI provide
consistent or complementary views through cross-source association,
imputation gain, extreme timing, and event detection. Tasks~2a--2d then
ask whether these source differences are learnable in operational
settings: cloud-gap LST-UHI imputation, station-sparse AirT-UHI
imputation, 1--96~h forecasting, and driver attribution after
removing location-, hour-, and month-specific climatology. Finally,
Task~3 evaluates zero-shot transfer, motivated by incomplete satellite
and station records in many cities. Following cross-city urban computing
and Local Climate Zone evidence~\cite{ref21,ref22,ref55} -- including empirical
tests showing that Local Climate Zone classifiers transfer conditionally
across cities, with success governed by morphological and ecoregion
similarity rather than geographic proximity~\cite{ref55} -- Task~3a compares
climate-homogeneous and climate-diverse source sets under matched size,
while Task~3b tests directed transfer asymmetry among seven cities.
Because UHI depends on climate, morphology, radiation, moisture, heat
storage, wind, and boundary-layer mixing~\cite{ref10,ref12,ref17,ref18,ref23}, transfer is evaluated
by both climate labels and target-state coverage.

We compare statistical/geostatistical, classical machine learning, deep
spatiotemporal, and time-series foundation-model baselines. Throughout
the experiments, base denotes UHI-history-only inputs; +m, +s, and
+s+m add meteorological drivers, static urban morphology, or both.
Tables~\ref{tab:baseline-groups} and \ref{tab:task-overview} summarize
the baseline taxonomy, input configurations, task protocols, city
coverage, and metrics.

\subsection{Evaluation Metrics by Task}
\label{sec:metrics-by-task}

For a set of $N$ predictions $\hat{y}_i$ against targets $y_i$\footnote{
$\hat{y}_i$ and $y_i$ are prediction and target for sample $i$; $N$ is
the sample count.}:
\[
\text{MAE} = \frac{1}{N}\sum_{i=1}^{N} |\hat{y}_i - y_i|, \qquad
\text{RMSE} = \sqrt{\frac{1}{N}\sum_{i=1}^{N} (\hat{y}_i - y_i)^2}.
\]
For binary extreme-event detection with true positives $TP$, false
positives $FP$, and false negatives $FN$\footnote{$TP$, $FP$, and $FN$
are true-positive, false-positive, and false-negative counts.}:
\[
\begin{aligned}
F1 &= \frac{2\,TP}{2\,TP + FP + FN},\\
\text{MissRate} &= \frac{FN}{FN + TP}, \qquad
\text{FAR} = \frac{FP}{FP + TP}.
\end{aligned}
\]
For Analysis~1b, let $\mathcal{E}_{c,s}$ be the set of extreme hours for
city $c$ and source $s$ after applying the source-specific threshold and
duration rule, and let $h(t)\in\{0,\ldots,23\}$ be the local hour of
timestamp $t$\footnote{$\mathcal{E}_{c,s}$ is the extreme-hour set for
city $c$ and source $s$; $h(t)$ is local hour; $\mathcal{H}_{\mathrm{night}}$
is the nighttime local-hour set.}. The share of extreme hours assigned to local hour $h$ is
\[
q_{c,s}(h) =
\frac{|\{t\in\mathcal{E}_{c,s}: h(t)=h\}|}{|\mathcal{E}_{c,s}|},
\qquad
\sum_{h=0}^{23} q_{c,s}(h)=1.
\]
The night-extreme fraction is the share of extreme hours that fall in
the predefined nighttime local-hour set $\mathcal{H}_{\mathrm{night}}$:
\[
\text{NightFrac}_{c,s} =
\frac{|\{t\in\mathcal{E}_{c,s}: h(t)\in\mathcal{H}_{\mathrm{night}}\}|}
{|\mathcal{E}_{c,s}|}.
\]
The peak-hour timing is the local hour with the largest extreme-hour
share:
\[
h^{\star}_{c,s} = \arg\max_{h\in\{0,\ldots,23\}} q_{c,s}(h).
\]
For covariate-gain analysis, comparing a model trained with an
additional feature block against its base-input counterpart\footnote{
$\text{MAE}_{\text{base}}$ and $\text{MAE}_{+\text{feature}}$ are MAE
before and after adding the feature block.}:
\[
\Delta\text{MAE} = \text{MAE}_{+\text{feature}} - \text{MAE}_{\text{base}},
\]
where negative $\Delta\text{MAE}$ indicates improvement.

\begin{table*}[!htbp]
  \caption{Baseline groups used in UHI-Bench.}
  \label{tab:baseline-groups}
  \small
  \setlength{\tabcolsep}{4pt}
  \renewcommand{\arraystretch}{1.16}
\begin{tabular}{p{3.2cm} p{6.2cm} p{4.2cm} p{3.1cm}}
    \toprule
    Group & Baselines & Main role & Input configurations \\
    \midrule
    Statistical / geostatistical &
    Persistence, climatology, percentile rule, seasonal naive,
    inverse-distance weighting, ordinary kriging, universal kriging,
    regression kriging &
    Simple temporal and spatial references for forecasting, event
    detection, and imputation &
    base; +s where applicable \\
    \addlinespace
    Classical machine learning &
    RF, XGBoost, Isolation Forest, One-Class SVM, linear
    models, regularized regressions &
    Covariate-aware imputation, detection, attribution, forecasting,
    and transfer analysis &
    base, +m, +s, +s+m when supported \\
    \addlinespace
    Deep spatiotemporal models &
    SAITS, GNN/IGNNK-style models, GraphWaveNet, LSTM, PatchTST-style
    models, DLinear &
    Temporal, spatial, and graph-based inductive biases for imputation
    and forecasting &
    base, +m, +s, +s+m depending on model interface \\
    \addlinespace
    Time-series foundation models &
    TimesFM, Chronos-2, MOIRAI-2, ChronosAdapter &
    Zero-shot or low-adaptation sequence modeling under missingness,
    sparse observations, forecasting, and cross-city transfer &
    TimesFM: base only; Chronos-2, MOIRAI-2, and ChronosAdapter: base,
    +m, +s, +s+m where supported \\
    \bottomrule
  \end{tabular}
\end{table*}
\begin{table*}[!htbp]
  \caption{Complete task evaluation protocols, city coverage, and metrics used in the benchmark experiments.}
  \label{tab:task-overview}
  \small
  \setlength{\tabcolsep}{2pt}
  \renewcommand{\arraystretch}{1.16}
  \centering
  \begin{tabular}{@{}p{0.15\textwidth}p{0.41\textwidth}p{0.29\textwidth}p{0.08\textwidth}@{}}
    \toprule
    Task & Evaluation protocol & Cities / source-target sets & Metrics \\
    \midrule
    Analysis 1a: Cross-source consistency &
    Quantify LST-UHI/AirT-UHI consistency via temporal, spatial-pattern,
    and lagged correlation, plus cross-source imputation gains. &
    16 dual-source cities &
    Pearson ($r$); $\Delta$MAE \\
    Analysis 1b: Heat-risk timing diagnostics &
    Identify source-specific extreme periods in city-mean hourly UHI using
    training-period P95 thresholds (min.\ 3 consecutive hours). &
    Berlin, Munich, Hamburg, Cairo, Lagos, Johannesburg, Riyadh &
    Local-hour share; night fraction; peak hour \\
    Task 1: Extreme-event detection &
    Detect source-specific extreme events under base, +m, +s, and
    +s+m inputs. &
    Berlin, Munich, Hamburg, Cairo, Lagos, Johannesburg, Riyadh &
    F1 (main); FAR \\
    Task 2a: LST cloud-gap imputation &
    Reconstruct LST-UHI using near-clear references and real cloud masks;
    errors computed over cloud-covered pixels across four coverage bins
    (0--25/25--50/50--75/$>$75\%). Evaluate base, +m, +s,
    +s+m where supported. &
    Cairo, Bucharest, Lagos &
    MAE (main); RMSE (supp.) \\
    Task 2b: AirT sparse imputation &
    Reconstruct AirT-UHI from simulated station-like masks under the same
    four missing-rate bins. Evaluate base, +m, +s, +s+m
    where supported. &
    Lagos, Cologne, Riyadh, Munich, Cairo &
    MAE (main); RMSE (supp.) \\
    Task 2c: Dual-source UHI forecasting &
    Forecast both UHI sources at 1/6/12/24/48/96~h horizons, comparing
    base, +m, +s, and +s+m inputs. &
    Munich, Berlin, Cologne, Lagos, Johannesburg, Riyadh &
    MAE (main); RMSE (supp.) \\
    Task 2d: Driver attribution &
    Attribute baseline-removed UHI anomalies via eight model-family
    settings with feature-block ablations. &
    16 dual-source cities &
    SHAP; rank stability \\
\bottomrule
  \end{tabular}
\end{table*}
\begin{table*}[!htbp]
  \caption{Complete Task~1 extreme-event detection results (F1, higher
  is better) for Berlin, Munich, Hamburg, Cairo, Lagos, Johannesburg, and
  Riyadh. The
  table adds the +s and +s+m F1 rows for the classical, unsupervised,
  and LSTM baselines; suffixes indicate input configuration.}
  \label{tab:1c-fm-complete}
  \small
  \setlength{\tabcolsep}{1pt}
  \centering
  \resizebox{0.92\textwidth}{!}{%
  \begin{tabular}{l l c c c c c c c c c c c c c c}
    \toprule
    & & \multicolumn{14}{c}{F1} \\
    \cmidrule(lr){3-16}
    & & \multicolumn{2}{c}{\textcolor{cfblue}{Berlin (Cfb)}} &
    \multicolumn{2}{c}{\textcolor{cfblue}{Munich (Cfb)}} &
    \multicolumn{2}{c}{\textcolor{cfblue}{Hamburg (Cfb)}} &
    \multicolumn{2}{c}{\textcolor{cfblue}{Cairo (BWh)}} &
    \multicolumn{2}{c}{\textcolor{cfblue}{Lagos (Aw)}} &
    \multicolumn{2}{c}{\textcolor{cfblue}{Johannesburg (Cwb)}} &
    \multicolumn{2}{c}{\textcolor{cfblue}{Riyadh (BWh)}} \\
    \cmidrule(lr){3-4}\cmidrule(lr){5-6}\cmidrule(lr){7-8}\cmidrule(lr){9-10}\cmidrule(lr){11-12}\cmidrule(lr){13-14}\cmidrule(lr){15-16}
    Group & Model & AirT-UHI & LST-UHI & AirT-UHI & LST-UHI & AirT-UHI &
    LST-UHI & AirT-UHI & LST-UHI & AirT-UHI & LST-UHI & AirT-UHI & LST-UHI &
    AirT-UHI & LST-UHI \\
    \midrule
    Stat/Geo & Percentile & \vbase{0.894} & \vbase{0.197} & \vbase{0.883} & \vbase{0.020} & \vbase{0.891} & \vbase{0.054} & \vbase{0.774} & \vbase{0.511} & \vbase{0.654} & \vbase{0.248} & \vbase{0.760} & \vbase{0.621} & \vbase{0.786} & \vbase{0.329} \\
    Stat/Geo & SeasonalNaive & \vbase{0.548} & \vbase{0.182} & \vbase{0.538} & \vbase{0.019} & \vbase{0.575} & \vbase{0.049} & \vbase{0.406} & \vbase{0.224} & \vbase{0.371} & \vbase{0.249} & \vbase{0.508} & \vbase{0.540} & \vbase{0.462} & \vbase{0.187} \\
    \midrule
    ML & RF & \vbase{0.956} & \vbase{0.322} & \vbase{0.949} & \vbase{0.017} & \vbase{0.951} & \vbase{0.032} & \vbase{0.850} & \vbase{0.579} & \vbase{0.740} & \vbase{0.378} & \vbase{0.739} & \vbase{0.663} & \vbase{0.870} & \vbase{0.481} \\
    ML & RF+m & \vmet{0.941} & \vmet{0.313} & \vmet{0.941} & \vmet{0.000} & \vmet{0.932} & \cfgbest\vmetB{0.127} & \vmetU{0.869} & \vmet{0.581} & \vmet{0.752} & \vmet{0.396} & \vmet{0.753} & \vmet{0.659} & \vmet{0.872} & \vmet{0.506} \\
\bottomrule
  \end{tabular}%
  }
\end{table*}
\begin{table*}[!htbp]
  \caption{Additional Task~2a LST cloud-gap imputation results for baselines
  omitted from the compact main table. Each cell reports MAE. Method suffixes
  indicate input configuration. UnivKrig = Universal Kriging; GWNet =
  GraphWaveNet.}
  \label{tab:task2a-remaining}
  \small
  \setlength{\tabcolsep}{4pt}
  \centering
  \begin{tabular}{l l c c c c c c c c c c c c}
    \toprule
    & & \multicolumn{12}{c}{MAE} \\
    \cmidrule(lr){3-14}
    & & \multicolumn{4}{c}{Cairo} & \multicolumn{4}{c}{Bucharest} &
    \multicolumn{4}{c}{Lagos} \\
    \cmidrule(lr){3-6}\cmidrule(lr){7-10}\cmidrule(lr){11-14}
    Group & Model & 0--25\% & 25--50\% & 50--75\% & $>$75\% & 0--25\% &
    25--50\% & 50--75\% & $>$75\% & 0--25\% & 25--50\% & 50--75\% & $>$75\% \\
    \midrule
    Stat/Geo & UnivKrig & \vbase{1.318} & \vbase{1.435} & \vbase{1.829} & \vbase{2.343} & \vbase{1.283} & \vbase{1.539} & \vbase{1.650} & \vbase{2.101} & \vbase{1.616} & \vbase{1.626} & \vbase{1.919} & \vbase{2.185} \\
    Stat/Geo & IDW\_static & \vstat{1.405} & \vstat{1.568} & \vstat{2.082} & \vstat{2.258} & \vstat{0.989} & \vstat{1.379} & \vstat{1.285} & \vstat{1.440} & \vstat{1.500} & \vstat{1.656} & \vstat{2.005} & \vstat{2.521} \\
    \midrule
    ML & RF & \vbase{1.639} & \vbase{1.682} & \vbase{2.390} & \vbase{2.527} & \vbase{0.964} & \vbase{1.403} & \vbase{1.344} & \vbase{1.514} & \vbase{1.554} & \vbase{1.779} & \vbase{2.079} & \vbase{2.511} \\
    ML & XGBoost & \vbase{1.691} & \vbase{1.710} & \vbase{2.227} & \vbase{2.588} & \vbase{1.006} & \vbase{1.391} & \vbase{1.330} & \vbase{1.503} & \vbase{1.571} & \vbase{1.805} & \vbase{2.133} & \vbase{2.579} \\
    ML & RF+s & \vstat{1.178} & \vstat{1.194} & \vstat{1.527} & \vstat{1.733} & \vstat{0.955} & \vstat{1.257} & \vstat{1.208} & \vstat{1.355} & \vstat{1.041} & \vstat{1.121} & \vstat{1.237} & \vstat{1.387} \\
\bottomrule
  \end{tabular}
\end{table*}
\begin{table*}[!htbp]
  \caption{Continuation of Table~\ref{tab:1b}: Task~2b AirT station-sparse
  imputation MAE for additional Stat/Geo, ML covariate, and FM baselines.
  Each cell reports MAE by station-missing bin for Lagos, Cologne, and
  Riyadh. OrdKrig/RegKrig/UnivKrig = Ordinary/Regression/Universal Kriging.}
  \label{tab:task2b-table4-continuation}
  \small
  \setlength{\tabcolsep}{1pt}
  \centering
  \resizebox{0.72\textwidth}{!}{%
  \begin{tabular}{l l c c c c c c c c c c c c}
    \toprule
    & & \multicolumn{12}{c}{MAE, by station missing-\% bin} \\
    \cmidrule(lr){3-14}
    & & \multicolumn{4}{c}{\textcolor{cfblue}{Lagos (Aw)}} & \multicolumn{4}{c}{\textcolor{cfblue}{Cologne (Cfb)}} &
    \multicolumn{4}{c}{\textcolor{cfblue}{Riyadh (BWh)}} \\
    \cmidrule(lr){3-6}\cmidrule(lr){7-10}\cmidrule(lr){11-14}
    Group & Method & 0--25\% & 25--50\% & 50--75\% & $>$75\% & 0--25\% &
    25--50\% & 50--75\% & $>$75\% & 0--25\% & 25--50\% & 50--75\% & $>$75\% \\
    \midrule
    Stat/Geo & IDW & \vbase{0.040} & \vbase{0.043} & \vbase{0.048} &
    \vbase{0.059} & \vbase{0.0992} & \vbase{0.1055} & \vbase{0.1141} &
    \vbase{0.1370} & \vbase{0.0585} & \vbase{0.0617} & \vbase{0.0671} &
    \vbase{0.0817} \\
    Stat/Geo & OrdKrig & \vbase{0.055} & \vbase{0.064} &
    \vbase{0.081} & \vbase{0.125} & \vbase{0.1772} & \vbase{0.1975} &
    \vbase{0.2266} & \vbase{0.2559} & \vbase{0.0708} & \vbase{0.0794} &
    \vbase{0.0958} & \vbase{0.1525} \\
    Stat/Geo & UnivKrig & \vbase{0.069} & \vbase{0.072} &
    \vbase{0.071} & \vbase{0.081} & \vbase{0.1288} & \vbase{0.1313} &
    \vbase{0.1399} & \vbase{0.1729} & \vbase{0.1042} & \vbase{0.1060} &
    \vbase{0.1051} & \vbase{0.1126} \\
    Stat/Geo & RegKrig+s & \vstat{0.037} & \vstat{0.038} &
    \vstat{0.037} & \vstat{0.042} & \vstat{0.1020} & \vstat{0.1068} &
    \vstat{0.1112} & \vstat{0.1260} & \vstat{0.0666} & \vstat{0.0668} &
    \vstat{0.0638} & \vstat{0.0719} \\
    \midrule
    ML & RF+m & \vmet{0.0513} & \vmet{0.0518} & \vmet{0.0532} &
    \vmet{0.0665} & \vmet{0.1102} & \vmet{0.1176} & \vmet{0.1285} &
    \vmet{0.1565} & \vmet{0.0765} & \vmet{0.0760} & \vmet{0.0778} &
    \vmet{0.0925} \\
\bottomrule
  \end{tabular}%
  }
  \par\vspace{5pt}
  \normalsize
  \caption{Complete Task~2b station-sparse AirT imputation results for
  Munich and Cairo. Each cell reports MAE by station-missing bin. Method
  suffixes indicate input configuration. Within each city/bin, the overall best MAE is
  bolded, and the second best is underlined. OrdKrig/RegKrig/UnivKrig =
  Ordinary/Regression/Universal Kriging; GWNet = GraphWaveNet; IGNNK = GNN (IGNNK).}
  \label{tab:task2b-complete}
  \small
  \setlength{\tabcolsep}{1pt}
  \centering
  \resizebox{0.53\textwidth}{!}{%
  \begin{tabular}{l l c c c c c c c c}
    \toprule
    & & \multicolumn{8}{c}{MAE} \\
    \cmidrule(lr){3-10}
    & & \multicolumn{4}{c}{Munich} & \multicolumn{4}{c}{Cairo} \\
    \cmidrule(lr){3-6}\cmidrule(lr){7-10}
    Group & Method & 0--25\% & 25--50\% & 50--75\% & $>$75\% & 0--25\% &
    25--50\% & 50--75\% & $>$75\% \\
    \midrule
    Stat/Geo & IDW & \vbase{0.0269} & \vbase{0.0274} & \vbase{0.0297} & \vbase{0.0368} & \vbase{0.0999} & \vbase{0.1020} & \vbase{0.1095} & \vbase{0.1252} \\
    Stat/Geo & OrdKrig & \vbase{0.0405} & \vbase{0.0438} & \vbase{0.0558} & \vbase{0.0803} & \vbase{0.1212} & \vbase{0.1294} & \vbase{0.1534} & \vbase{0.2188} \\
    Stat/Geo & UnivKrig & \vbase{0.0501} & \vbase{0.0489} & \vbase{0.0499} & \vbase{0.0625} & \vbase{0.1738} & \vbase{0.1715} & \vbase{0.1629} & \vbase{0.1627} \\
    ML & RF & \vbase{0.0377} & \vbase{0.0360} & \vbase{0.0376} & \vbase{0.0480} & \vbase{0.1237} & \vbase{0.1208} & \vbase{0.1219} & \vbase{0.1391} \\
    ML & XGBoost & \vbase{0.0356} & \vbase{0.0344} & \vbase{0.0356} & \vbase{0.0479} & \vbase{0.1275} & \vbase{0.1245} & \vbase{0.1257} & \vbase{0.1444} \\
\bottomrule
  \end{tabular}%
  }
\end{table*}

\clearpage
\onecolumn
\input{tab_1d_fm_complete.tex}

\clearpage
\twocolumn

\begin{figure*}[!htbp]
  \centering
  \includegraphics[width=0.88\textwidth]{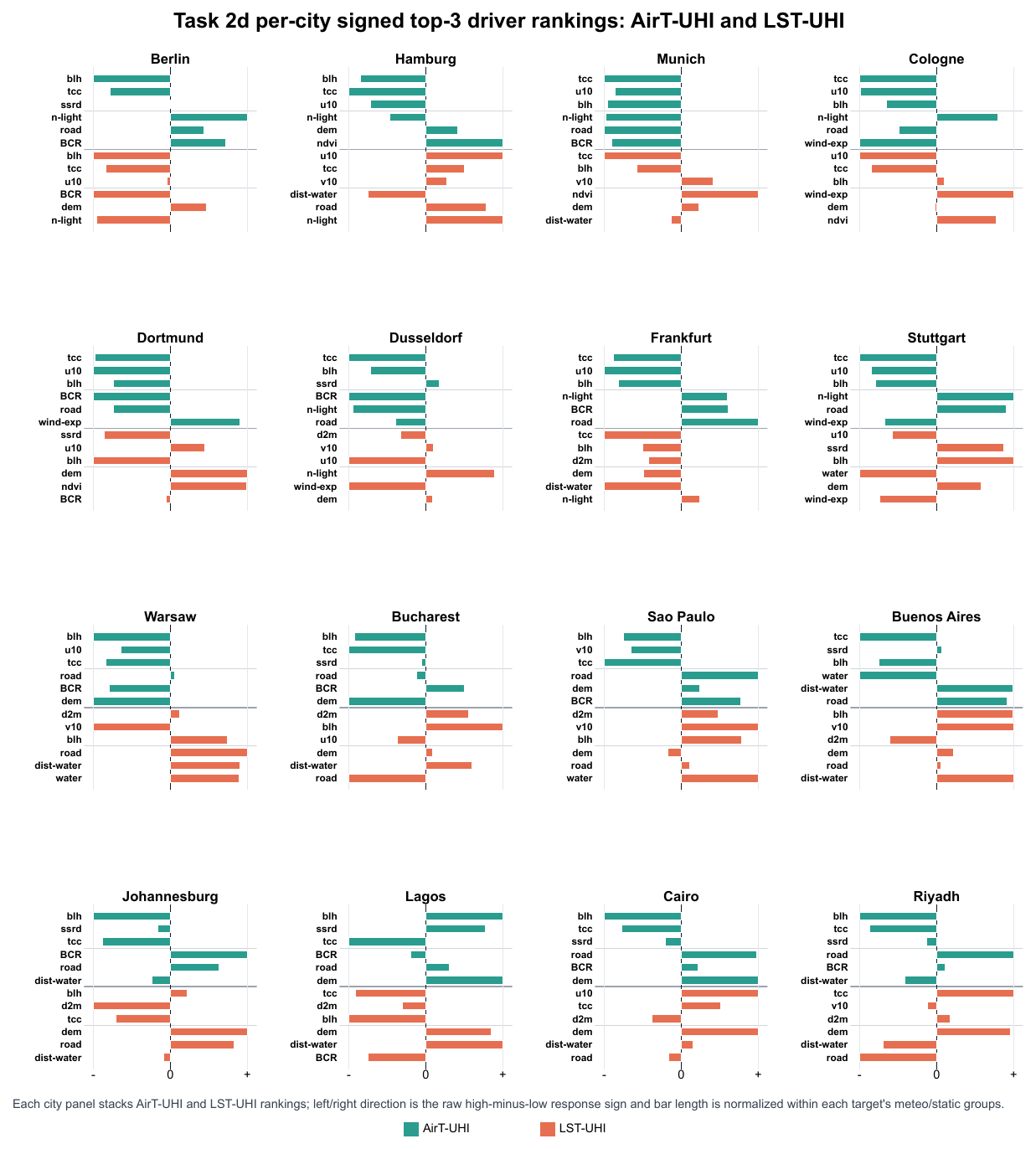}
  \caption{Task~2d per-city signed top-3 driver rankings for AirT-UHI and
  LST-UHI in one view. Each city panel stacks the XGBoost top-3
  meteorological drivers and top-3 static features for both targets; green
  bars denote AirT-UHI and orange bars denote LST-UHI. Bar direction is the
  raw high-minus-low response sign; bar length is normalized separately within
  each target's meteorological and static groups, so that within-city ranking
  patterns remain visible.}
  \label{fig:task2d-city-top3-bars-combined}
  \Description{Sixteen small horizontal bar charts showing each city's signed
  top-three meteorological and static driver rankings for both AirT-UHI and
  LST-UHI.}
\end{figure*}
\end{document}

%% file: appendix_dataset_summary_from_source.tex
\begin{longtable}{%
  >{\centering\arraybackslash}p{1.35cm}
  >{\centering\arraybackslash}p{2.75cm}
  >{\raggedright\arraybackslash}p{2.55cm}
  >{\centering\arraybackslash}p{2.05cm}
  >{\centering\arraybackslash}p{3.55cm}
  >{\centering\arraybackslash}p{1.45cm}
  >{\centering\arraybackslash}p{1.35cm}
}
\caption{Full Dataset Summary for UHI-Bench}
\label{tab:dataset_summary_full} \\
\toprule
\textbf{Data} &
\textbf{Domain} &
\textbf{Spatial} &
\textbf{Number of} &
\textbf{Temporal} &
\textbf{\# Temporal} &
\textbf{Missing} \\
\textbf{Format} &
\textbf{Type} &
\textbf{Region} &
\begin{tabular}[c]{@{}c@{}}\textbf{Grids/}\\\textbf{Stations}\end{tabular} &
\textbf{Range} &
\textbf{Step} &
\textbf{Rate} \\
\midrule
\endfirsthead

\caption{Full Dataset Summary for UHI-Bench} \\
\toprule
\textbf{Data} &
\textbf{Domain} &
\textbf{Spatial} &
\textbf{Number of} &
\textbf{Temporal} &
\textbf{\# Temporal} &
\textbf{Missing} \\
\textbf{Format} &
\textbf{Type} &
\textbf{Region} &
\begin{tabular}[c]{@{}c@{}}\textbf{Grids/}\\\textbf{Stations}\end{tabular} &
\textbf{Range} &
\textbf{Step} &
\textbf{Rate} \\
\midrule
\endhead

\bottomrule
\endlastfoot

Grid
& Air Temperature UHI\textsuperscript{a}
& Berlin, DE
& 2{,}817
& 2015/01/01 -- 2025/12/31
& 96{,}432
& 0.00\% \\

&
&
Hamburg, DE
& 2{,}216
& 2015/01/01 -- 2025/12/31
& 96{,}432
& 0.00\% \\

&
&
Munich, DE
& 4{,}203
& 2015/01/01 -- 2025/12/31
& 96{,}432
& 0.00\% \\

&
&
Cologne, DE
& 1{,}148
& 2015/01/01 -- 2025/12/31
& 96{,}432
& 0.00\% \\

&
&
Dortmund, DE
& 864
& 2015/01/01 -- 2025/12/31
& 96{,}432
& 0.00\% \\

&
&
D\"{u}sseldorf, DE
& 544
& 2015/01/01 -- 2025/12/31
& 96{,}432
& 0.00\% \\

&
&
Frankfurt, DE
& 1{,}632
& 2015/01/01 -- 2025/12/31
& 96{,}432
& 0.00\% \\

&
&
Stuttgart, DE
& 1{,}144
& 2015/01/01 -- 2025/12/31
& 96{,}432
& 0.00\% \\

& Air Temperature UHI (downscaled)\textsuperscript{b}
& Cairo, EG
& 6{,}156
& 2015/01/01 -- 2025/12/31
& 96{,}432
& $\approx$0\% \\

&
&
Johannesburg, ZA
& 5{,}776
& 2015/01/01 -- 2025/12/31
& 96{,}432
& $\approx$0\% \\

&
&
Lagos, NG
& 5{,}751
& 2015/01/01 -- 2025/12/31
& 96{,}432
& $\approx$0\% \\

&
&
Riyadh, SA
& 5{,}396
& 2015/01/01 -- 2025/12/31
& 96{,}432
& $\approx$0\% \\

&
&
Bucharest, RO
& 5{,}041
& 2015/01/01 -- 2025/12/31
& 96{,}432
& $\approx$0\% \\

&
&
Warsaw, PL
& 5{,}396
& 2015/01/01 -- 2025/12/31
& 96{,}432
& $\approx$0\% \\

&
&
S\~{a}o Paulo, BR
& 6{,}536
& 2015/01/01 -- 2025/12/31
& 96{,}432
& $\approx$0\% \\

&
&
Buenos Aires, AR
& 5{,}751
& 2015/01/01 -- 2025/12/31
& 96{,}432
& $\approx$0\% \\

\midrule

Grid
& LST-UHI\textsuperscript{c}
& Berlin, DE
& 3{,}886
& 2015/01/01 -- 2025/12/31
& 96{,}432
& 74.6\% \\

&
&
Hamburg, DE
& 3{,}098
& 2015/01/01 -- 2025/12/31
& 96{,}432
& 76.3\% \\

&
&
Munich, DE
& 5{,}770
& 2015/01/01 -- 2025/12/31
& 96{,}432
& 73.7\% \\

&
&
Cologne, DE
& 1{,}148
& 2015/01/01 -- 2025/12/31
& 96{,}432
& 75.6\% \\

&
&
Dortmund, DE
& 864
& 2015/01/01 -- 2025/12/31
& 96{,}432
& 75.7\% \\

&
&
D\"{u}sseldorf, DE
& 544
& 2015/01/01 -- 2025/12/31
& 96{,}432
& 75.2\% \\

&
&
Frankfurt, DE
& 1{,}632
& 2015/01/01 -- 2025/12/31
& 96{,}432
& 74.9\% \\

&
&
Stuttgart, DE
& 1{,}144
& 2015/01/01 -- 2025/12/31
& 96{,}432
& 72.9\% \\

&
&
Cairo, EG
& 6{,}156
& 2015/01/01 -- 2025/12/31
& 96{,}432
& 20.0\% \\

&
&
Johannesburg, ZA
& 5{,}776
& 2015/01/01 -- 2025/12/31
& 96{,}432
& 36.4\% \\

&
&
Lagos, NG
& 5{,}751
& 2015/01/01 -- 2025/12/31
& 96{,}432
& 75.6\% \\

&
&
Riyadh, SA
& 5{,}396
& 2015/01/01 -- 2025/12/31
& 96{,}432
& 22.7\% \\

&
&
Bucharest, RO
& 5{,}041
& 2015/01/01 -- 2025/12/31
& 96{,}432
& 57.2\% \\

&
&
Warsaw, PL
& 5{,}396
& 2015/01/01 -- 2025/12/31
& 96{,}432
& 69.9\% \\

&
&
S\~{a}o Paulo, BR
& 6{,}536
& 2015/01/01 -- 2025/12/31
& 96{,}432
& 67.3\% \\

&
&
Buenos Aires, AR
& 5{,}751
& 2015/01/01 -- 2025/12/31
& 96{,}432
& 74.9\% \\

\midrule

Feature
& Static: BCR\textsuperscript{d}, road density\textsuperscript{e}, POI density\textsuperscript{e}, water ratio\textsuperscript{f}, distance to water\textsuperscript{e}, mean height\textsuperscript{g}, DEM\textsuperscript{h}, wind exposure proxy\textsuperscript{i}
& Berlin, DE
& 2{,}817
& Static
& 1
& $\approx$0\% \\

&
&
Hamburg, DE
& 2{,}230
& Static
& 1
& $\approx$0\% \\

&
&
Munich, DE
& 4{,}203
& Static
& 1
& $\approx$0\% \\

&
&
Cologne, DE
& 1{,}148
& Static
& 1
& $\approx$0\% \\

&
&
Dortmund, DE
& 864
& Static
& 1
& $\approx$0\% \\

&
&
D\"{u}sseldorf, DE
& 544
& Static
& 1
& $\approx$0\% \\

&
&
Frankfurt, DE
& 1{,}632
& Static
& 1
& $\approx$0\% \\

&
&
Stuttgart, DE
& 1{,}144
& Static
& 1
& $\approx$0\% \\

&
&
Cairo, EG
& 6{,}156
& Static
& 1
& $\approx$0\% \\

&
&
Johannesburg, ZA
& 5{,}776
& Static
& 1
& $\approx$0\% \\

&
&
Lagos, NG
& 5{,}751
& Static
& 1
& $\approx$0\% \\

&
&
Riyadh, SA
& 5{,}396
& Static
& 1
& $\approx$0\% \\

&
&
Bucharest, RO
& 5{,}041
& Static
& 1
& $\approx$0\% \\

&
&
Warsaw, PL
& 5{,}396
& Static
& 1
& $\approx$0\% \\

&
&
S\~{a}o Paulo, BR
& 6{,}536
& Static
& 1
& $\approx$0\% \\

&
&
Buenos Aires, AR
& 5{,}751
& Static
& 1
& $\approx$0\% \\

\midrule

Feature
& Annual: VIIRS nightlight\textsuperscript{j}, Sentinel-2 NDVI\textsuperscript{k}, ERA5-Land wind speed\textsuperscript{i}
& Berlin, DE
& 2{,}817
& 2015--2025
& 11
& $\approx$0\% \\

&
&
Hamburg, DE
& 2{,}230
& 2015--2025
& 11
& $\approx$0\% \\

&
&
Munich, DE
& 4{,}203
& 2015--2025
& 11
& $\approx$0\% \\

&
&
Cologne, DE
& 1{,}148
& 2015--2025
& 11
& $\approx$0\% \\

&
&
Dortmund, DE
& 864
& 2015--2025
& 11
& $\approx$0\% \\

&
&
D\"{u}sseldorf, DE
& 544
& 2015--2025
& 11
& $\approx$0\% \\

&
&
Frankfurt, DE
& 1{,}632
& 2015--2025
& 11
& $\approx$0\% \\

&
&
Stuttgart, DE
& 1{,}144
& 2015--2025
& 11
& $\approx$0\% \\

&
&
Cairo, EG
& 6{,}156
& 2015--2025
& 11
& $\approx$0\% \\

&
&
Johannesburg, ZA
& 5{,}776
& 2015--2025
& 11
& $\approx$0\% \\

&
&
Lagos, NG
& 5{,}751
& 2015--2025
& 11
& $\approx$0\% \\

&
&
Riyadh, SA
& 5{,}396
& 2015--2025
& 11
& $\approx$0\% \\

&
&
Bucharest, RO
& 5{,}041
& 2015--2025
& 11
& $\approx$0\% \\

&
&
Warsaw, PL
& 5{,}396
& 2015--2025
& 11
& $\approx$0\% \\

&
&
S\~{a}o Paulo, BR
& 6{,}536
& 2015--2025
& 11
& $\approx$0\% \\

&
&
Buenos Aires, AR
& 5{,}751
& 2015--2025
& 11
& $\approx$0\% \\

\midrule

Grid
& ERA5/ERA5-Land: $u_{10}$, $v_{10}$, total cloud cover, 2-m dew point, boundary-layer height, surface solar radiation downwards\textsuperscript{l}
& Berlin, DE
& 2{,}817
& 2015/01/01 -- 2025/12/31
& 96{,}432
& $\approx$0\% \\

&
&
Hamburg, DE
& 2{,}230
& 2015/01/01 -- 2025/12/31
& 96{,}432
& $\approx$0\% \\

&
&
Munich, DE
& 4{,}203
& 2015/01/01 -- 2025/12/31
& 96{,}432
& $\approx$0\% \\

&
&
Dortmund, DE
& 864
& 2015/01/01 -- 2025/12/31
& 96{,}432
& $\approx$0\% \\

&
&
Stuttgart, DE
& 1{,}144
& 2015/01/01 -- 2025/12/31
& 96{,}432
& $\approx$0\% \\

&
&
D\"{u}sseldorf, DE
& 544
& 2015/01/01 -- 2025/12/31
& 96{,}432
& $\approx$0\% \\

&
&
Cologne, DE
& 1{,}148
& 2015/01/01 -- 2025/12/31
& 96{,}432
& $\approx$0\% \\

&
&
Frankfurt, DE
& 1{,}632
& 2015/01/01 -- 2025/12/31
& 96{,}432
& $\approx$0\% \\

&
&
Cairo, EG
& 6{,}156
& 2015/01/01 -- 2025/12/31
& 96{,}432
& $\approx$0\% \\

&
&
Johannesburg, ZA
& 5{,}776
& 2015/01/01 -- 2025/12/31
& 96{,}432
& $\approx$0\% \\

&
&
Lagos, NG
& 5{,}751
& 2015/01/01 -- 2025/12/31
& 96{,}432
& $\approx$0\% \\

&
&
Riyadh, SA
& 5{,}396
& 2015/01/01 -- 2025/12/31
& 96{,}432
& $\approx$0\% \\

&
&
Bucharest, RO
& 5{,}041
& 2015/01/01 -- 2025/12/31
& 96{,}432
& $\approx$0\% \\

&
&
Warsaw, PL
& 5{,}396
& 2015/01/01 -- 2025/12/31
& 96{,}432
& $\approx$0\% \\

&
&
S\~{a}o Paulo, BR
& 6{,}536
& 2015/01/01 -- 2025/12/31
& 96{,}432
& $\approx$0\% \\

&
&
Buenos Aires, AR
& 5{,}751
& 2015/01/01 -- 2025/12/31
& 96{,}432
& $\approx$0\% \\

\midrule

Station
& Air Temperature UHI\textsuperscript{m}
& Rome, IT
& 17 stations
& JJA 2019--2020
& 4{,}416
& 5.17\% \\

&
&
Temuco, CL
& 50 stations
& 2017/01/10 -- 2018/12/11
& 16{,}824
& 67.94\% \\

\midrule

Grid
& LST-UHI\textsuperscript{c}
& Tehran, IR
& 5{,}016
& 2019/01/01 -- 2025/12/31
& 61{,}368
& 42.0\% \\

&
&
Khartoum, SD
& 5{,}776
& 2019/01/01 -- 2025/12/31
& 61{,}368
& 29.7\% \\

&
&
Casablanca, MA
& 5{,}776
& 2019/01/01 -- 2025/12/31
& 61{,}368
& 61.0\% \\

&
&
Istanbul, TR
& 4{,}816
& 2019/01/01 -- 2025/12/31
& 61{,}368
& 71.6\% \\

\midrule

Feature
& Static: BCR\textsuperscript{d}, road density\textsuperscript{e}, POI density\textsuperscript{e}, water ratio\textsuperscript{f}, distance to water\textsuperscript{e}, mean height\textsuperscript{g}, DEM\textsuperscript{h}, wind exposure proxy\textsuperscript{i}
& Tehran, IR
& 5{,}016
& Static
& 1
& $\approx$0\% \\

&
&
Khartoum, SD
& 5{,}776
& Static
& 1
& $\approx$0\% \\

&
&
Casablanca, MA
& 5{,}776
& Static
& 1
& $\approx$0\% \\

&
&
Istanbul, TR
& 4{,}816
& Static
& 1
& $\approx$0\% \\

\midrule

Feature
& Annual: VIIRS nightlight\textsuperscript{j}, Sentinel-2 NDVI\textsuperscript{k}, ERA5-Land wind speed\textsuperscript{i}
& Tehran, IR
& 5{,}016
& 2019--2025
& 7
& $\approx$0\% \\

&
&
Khartoum, SD
& 5{,}776
& 2019--2025
& 7
& $\approx$0\% \\

&
&
Casablanca, MA
& 5{,}776
& 2019--2025
& 7
& $\approx$0\% \\

&
&
Istanbul, TR
& 4{,}816
& 2019--2025
& 7
& $\approx$0\% \\

\midrule

Grid
& ERA5/ERA5-Land: $u_{10}$, $v_{10}$, total cloud cover, 2-m dew point, boundary-layer height, surface solar radiation downwards\textsuperscript{l}
& Tehran, IR
& 5{,}016
& 2019/01/01 -- 2025/12/31
& 61{,}368
& $\approx$0\% \\

&
&
Khartoum, SD
& 5{,}776
& 2019/01/01 -- 2025/12/31
& 61{,}368
& $\approx$0\% \\

&
&
Casablanca, MA
& 5{,}776
& 2019/01/01 -- 2025/12/31
& 61{,}368
& $\approx$0\% \\

&
&
Istanbul, TR
& 4{,}816
& 2019/01/01 -- 2025/12/31
& 61{,}368
& $\approx$0\% \\

\end{longtable}
\noindent\parbox{\textwidth}{\scriptsize
\textsuperscript{a} German gridded AirT-UHI uses DWD HOSTRADA/UHI-MAP products.
\textsuperscript{b} International downscaled AirT-UHI uses ERA5 temperature with HOSTRADA-based residual correction.
\textsuperscript{c} LST-UHI is derived from MSG/SEVIRI LST from European Organisation for the Exploitation of Meteorological Satellites Satellite Application Facility on Land Surface Analysis.
\textsuperscript{d} BCR uses GHSL GHS-BUILT-S.
\textsuperscript{e} Road density, POI density, and distance to water use OpenStreetMap/Geofabrik extracts.
\textsuperscript{f} Water ratio uses JRC Global Surface Water.
\textsuperscript{g} Building-height features use GHSL GHS-BUILT-H and German-only GlobalBuildingAtlas LoD1 where available.
\textsuperscript{h} DEM uses Copernicus DEM GLO-30.
\textsuperscript{i} Wind exposure uses ERA5-Land wind components.
\textsuperscript{j} Nightlight uses NOAA/VIIRS DNB monthly composites.
\textsuperscript{k} NDVI uses Sentinel-2 Level-2A surface reflectance.
\textsuperscript{l} Hourly meteorology uses CDS ERA5/ERA5-Land products.
\textsuperscript{m} Station AirT-UHI uses ASTI-Network Rome and Temuco fixed/mobile station observations. Climate labels use the 1 km K\"oppen-Geiger map [53].}

%% file: tab_1d_fm_complete.tex
{\small
\setlength{\tabcolsep}{0.5pt}
\renewcommand{\arraystretch}{0.92}
\begin{longtable}{@{}l l l c c c c c c c@{}}
  \caption{Complete Task~2c forecasting results for Munich, Berlin, Cologne,
  Lagos, Johannesburg, and Riyadh. Each metric cell reports MAE,
  AirT-UHI/LST-UHI; Avg is the mean over the six displayed horizons.}
  \label{tab:1d-fm-complete}\\
  \toprule
  Metric & & & \multicolumn{7}{c}{MAE, AirT-UHI/LST-UHI} \\
  \cmidrule(lr){4-10}
  City & Group & Method & 1h & 6h & 12h & 24h & 48h & 96h & Avg \\
  \midrule
  \endfirsthead
  \caption[]{Complete Task~2c forecasting results for Munich, Berlin, Cologne, Lagos, Johannesburg, and Riyadh (continued). Each metric cell reports AirT-UHI/LST-UHI MAE.}\\
  \toprule
  Metric & & & \multicolumn{7}{c}{MAE, AirT-UHI/LST-UHI} \\
  \cmidrule(lr){4-10}
  City & Group & Method & 1h & 6h & 12h & 24h & 48h & 96h & Avg \\
  \midrule
  \endhead
    Munich & Stat/Geo & Persistence & \vbaseB{0.006}/\vbase{0.547} & \vbase{0.027}/\vbase{1.089} & \vbase{0.120}/\vbase{1.298} & \vbase{0.080}/\vbase{0.768} & \vbase{0.100}/\vbase{0.896} & \vbase{0.109}/\vbase{0.910} & \vbase{0.074}/\vbase{0.918} \\
    Munich & Stat/Geo & Climatology & \vbase{0.096}/\vbase{0.972} & \vbase{0.085}/\vbase{0.940} & \vbaseU{0.016}/\vbase{0.932} & \vbase{0.096}/\vbase{0.978} & \vbase{0.096}/\vbase{0.986} & \vbase{0.096}/\vbase{0.961} & \vbase{0.081}/\vbase{0.962} \\
    Munich & ML & XGBoost & \vbase{0.011}/\vbase{0.499} & \vbase{0.042}/\vbase{0.792} & \vbase{0.056}/\vbase{0.642} & \vbase{0.052}/\vbase{0.606} & \vbase{0.061}/\vbase{0.621} & \vbase{0.064}/\vbase{0.666} & \vbase{0.048}/\vbase{0.637} \\
    Munich & ML & XGBoost+m & \vmetU{0.009}/\vmet{0.474} & \vmetU{0.025}/\vmet{0.668} & \vmet{0.033}/\vmet{0.616} & \vmetU{0.039}/\vmet{0.628} & \vmetU{0.045}/\vmet{0.658} & \vmetU{0.049}/\vmet{0.655} & \vmetU{0.033}/\vmet{0.617} \\
    Munich & ML & XGBoost+s & \vstat{0.010}/\vstat{0.498} & \vstat{0.034}/\vstat{0.780} & \vstat{0.045}/\vstat{0.636} & \vstat{0.049}/\vstat{0.605} & \vstat{0.056}/\vstatU{0.612} & \vstat{0.061}/\vstat{0.672} & \vstat{0.042}/\vstat{0.634} \\
    Munich & ML & XGBoost+s+m & \vall{0.009}/\vall{0.472} & \vallB{0.024}/\vall{0.656} & \vall{0.031}/\vall{0.603} & \vallB{0.038}/\vall{0.640} & \vallB{0.044}/\vall{0.644} & \vallB{0.047}/\vall{0.687} & \vallB{0.032}/\vall{0.617} \\
    Munich & DL & LSTM & \vbase{0.017}/\vbase{0.463} & \vbase{0.033}/\vbase{0.581} & \vbase{0.017}/\vbase{0.607} & \vbase{0.082}/\vbaseB{0.575} & \vbase{0.092}/\vbase{0.627} & \vbase{0.094}/\vbase{0.649} & \vbase{0.056}/\vbase{0.583} \\
    Munich & DL & DLinear & \vbase{0.013}/\vbaseU{0.459} & \vbase{0.038}/\vbase{0.571} & \vbaseB{0.014}/\vbaseB{0.586} & \vbase{0.083}/\vbase{0.587} & \vbase{0.091}/\vbase{0.613} & \vbase{0.091}/\vbaseB{0.606} & \vbase{0.055}/\vbaseB{0.570} \\
\bottomrule
\end{longtable}
}